\documentclass{article}

\usepackage[preprint]{Rethink_VLA_Init}

\usepackage[utf8]{inputenc} 
\usepackage[T1]{fontenc}    
\usepackage{hyperref}       
\usepackage{url}            
\usepackage{booktabs}       
\usepackage{amsmath}        
\usepackage{amsfonts}       
\usepackage{nicefrac}       
\usepackage{microtype}      
\usepackage[table]{xcolor}  
\usepackage{enumitem}       
\usepackage{multirow}       
\usepackage{array}          
\usepackage{graphicx}       
\usepackage{wrapfig}        
\definecolor{linkgreen}{RGB}{0,110,70}
\definecolor{linkred}{RGB}{150,35,35}
\hypersetup{
  colorlinks=true,
  linkcolor=linkred,
  citecolor=linkgreen,
  urlcolor=linkgreen
}
\definecolor{gaincolor}{RGB}{0,130,0}
\definecolor{losscolor}{RGB}{200,0,0}
\newcommand{\gain}[1]{{\footnotesize\textcolor{gaincolor}{(+#1)}}}
\newcommand{\loss}[1]{{\footnotesize\textcolor{losscolor}{(#1)}}}
\newcommand{\zdelta}{{\footnotesize\textcolor{gray}{($\pm 0.0$)}}}

\title{Rethinking VLM Representation for VLA Initialization}

\author{%
     \bf Weifeng Lin\textsuperscript{1}
  ~~ \bf Siyuan Huang\textsuperscript{4}
  ~~ \bf Hao Li\textsuperscript{4}
  ~~ \bf Tingwei Chen\textsuperscript{1}
  ~~ \bf Ruichuan An\textsuperscript{3}\vspace{0.1cm}\\
  \bf Xinyu Wei\textsuperscript{2}
  ~~ \bf Jianbo Liu\textsuperscript{4}
  ~~ \bf Hongsheng Li\textsuperscript{1}
  \\[0.2cm]
  \textsuperscript{1}CUHK ~~~
  \textsuperscript{2}PolyU ~~~
  \textsuperscript{3}Peking University ~~~
  \textsuperscript{4}ACE Robotics ~~~
}

\begin{document}

\maketitle

\begin{abstract}
Vision-Language-Action (VLA) models widely adopt pretrained Vision-Language Models (VLMs) as policy backbones, 
yet it remains unclear what kind of pretrained VLM representation is useful as a VLA initialization. 
In this paper, we study VLA initialization as a controlled representation-design problem along three axes: capability-level embodied VQA supervision, parameter-update strategy, and robot-data pretraining.
Our experiments show that the original pretrained VLM representation is a key source of action performance.
However, embodied VQA adaptation does not yield uniform gains: its benefit depends on downstream bottlenecks, and gains from different capability domains are not simply additive.
For update strategy, LoRA provides a more reliable initialization than Full Finetune, indicating that overly reshaping the pretrained representation can weaken VLA initialization.
Robot-data pretraining further improves VLA initialization, with the strongest variant obtained by staged LoRA-based training.
Together, these findings suggest that effective VLM-to-VLA adaptation should inject action-relevant embodied and robot-trajectory signals while preserving the pretrained VLM representation that remains useful for action learning. Code is available at: \url{https://github.com/AFeng-x/Rethink_VLA_Initialization}
\end{abstract}

\section{Introduction}
\label{sec:intro}

Vision-Language-Action (VLA) models have become a prominent paradigm for language-conditioned robot control. A common design initializes the policy backbone from a pretrained Vision-Language Model (VLM), allowing the policy to inherit visual-language representations and modeling structure from large-scale pretraining. Recent VLA systems have improved through stronger backbones, action modules, and robot-data training~\citep{brohan2023rt,kim24openvla,black2024pi_0,intelligence2025pi_,kim2025fine,wen2025dexvla,bjorck2025gr00t,geminirobotics}. However, the choice of initialization remains important but is still not well understood. This motivates a basic question: \emph{what kind of pretrained VLM representation makes a useful VLA initialization?}

Recent studies have clarified parts of this issue. VLM4VLA~\citep{zhang2026vlm4vla} shows that stronger VLMs and higher embodied-related understanding performance do not necessarily lead to better action policies, while VLASER~\citep{yang2025vlaser} identifies a visual domain gap between VLM adaptation and action policy learning.
However, they primarily evaluate coarse-grained factors in isolation, leaving open how finer-grained choices, such as capability-level VQA domains and their compositions, reshape the VLM representation and how these changes can be systematically guided toward better VLA initialization.
We therefore treat VLA initialization as a controlled representation-design problem along three axes: (1) which embodied VQA domains to inject and combine; (2) how strongly to update the pretrained representation; and (3) how to couple perception-side VQA adaptation with action-side robot-data pretraining.
Concretely, we organize embodied VQA into seven capability-oriented domains, compare Full Finetune with LoRA~\citep{hu2022lora}, and examine robot-data pretraining alone or together with VQA supervision.

\begin{figure}[t]
  \centering
  \includegraphics[width=0.99\linewidth]{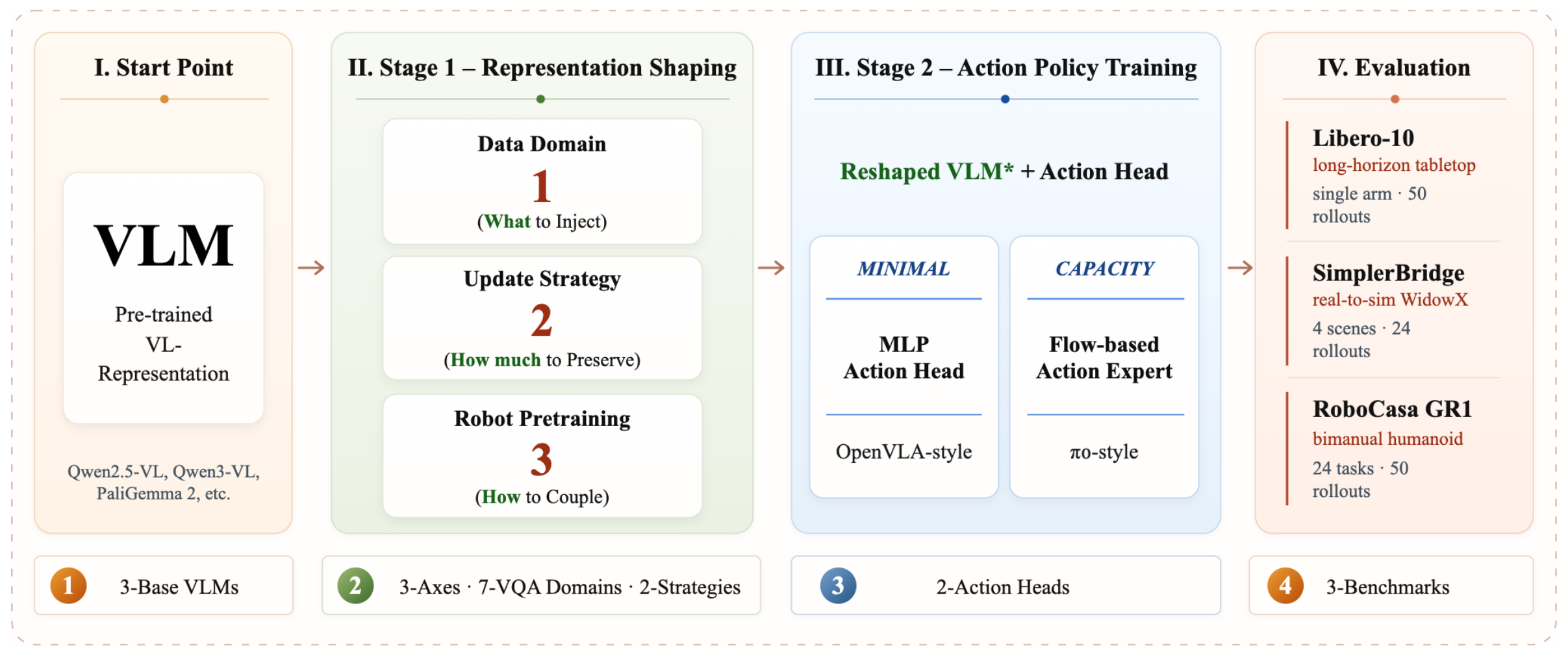}
  \caption{\textbf{Overview of our study design.} We evaluate the resulting initializations across multiple base VLMs, action heads, and simulated benchmarks.}
  \label{fig:framework}
  \vspace{-1mm}
\end{figure}

Our experiments reveal several significant and counterintuitive patterns. 
First, the original pretrained VLM representation is a major source of action performance, as policies trained from scratch drop by more than 20\% across all benchmarks.
Second, embodied VQA adaptation is useful conditionally: gains depend on the downstream bottleneck of the benchmark, and gains from different domains are not additive. The strongest improvement appears in a specific pairwise domain composition -- \{Grounding + Egocentric Understanding\}. This suggests that strengthening embodied capabilities is not a generally applicable recipe to improve VLA initialization.
Third, LoRA provides a more effective initialization than Full Finetune in VLM adaptation, indicating that action learning still benefits from the pretrained VLM. This effect also varies with VLM strength: across three VLMs with different strengths, LoRA gains shrink and Full Finetune degradation becomes more severe as the model becomes weaker. 
Finally, robot-data pretraining brings action-side supervision into the initialization process. Under the same downstream training recipe, it consistently improves VLA initialization, especially with LoRA-based updates. The best variant follows a staged route: adapt the VLM with Grounding and Egocentric Understanding, then continue with LoRA-based robot-data pretraining. This pattern suggests that even with action-side supervision, preserving the original VLM representation remains important for building a strong VLA initialization.

Overall, these results suggest a practical principle: effective VLA initialization requires injecting action-relevant signals while preserving the pretrained VLM representation that remains useful for action learning. The injected signal should match the downstream bottleneck, and the update strategy should avoid excessive representation drift. This reframes VLA initialization from a default backbone choice or data-scaling step into a controlled representation-design problem. We believe our findings provide practical guidance and insights for designing future VLM-to-VLA adaptation recipes.

\section{Related Work}
\label{sec:related}

\paragraph{VLM-Based VLA Architectures.}
Modern VLA systems often initialize action policies from pretrained VLMs, using their visual-language representations as priors for action learning. Representative systems include the RT series~\citep{brohan2022rt,brohan2023rt,o2023open}, OpenVLA-style models~\citep{kim24openvla,kim2025fine}, the $\pi$ series~\citep{black2024pi_0,intelligence2025pi_}, and recent generalist systems such as GR00T-N and Gemini Robotics~\citep{bjorck2025gr00t,geminirobotics}. Other recent work improves VLA performance through improved action modules and structure designs~\citep{pertsch2025fast,zheng2025x,cui2025openhelix,wen2025dexvla}. These works establish strong architectures, while the initialization mechanism remains less explicit. Our work instead studies what kind of VLM representation provides a better starting point for action learning.

\paragraph{Embodied VLM Adaptation.}
Another line of work adapts VLMs toward embodiment-relevant capabilities, including robotic reasoning and planning~\citep{driess2023palm,huang2022inner,mu2024embodiedgpt,RoboBrain2.0TechnicalReport,cosmosr1}, spatial and 3D understanding~\citep{chen2024spatialvlm,feng2025towards}, object affordance and referring~\citep{yuan2024robopoint,zhou2025roborefer}, and robot-oriented perception~\citep{sermanet2024robovqa,chen2025robo2vlmvisualquestionanswering}. These works motivate the intuition that stronger embodied understanding should improve VLA initialization, but it remains unclear which capability signals actually transfer to manipulation. We therefore use embodied VQA as a controlled adaptation signal and study its effect across capability domains and domain compositions.

\paragraph{VLM-to-VLA Transfer.}
Recent studies have begun to characterize transfer patterns from VLMs to downstream VLA policies.
VLM4VLA~\citep{zhang2026vlm4vla} studies how VLM choice and auxiliary embodied-task performance relate to downstream VLA performance, but leaves open how capability domains should be selected and combined as initialization variables.
VLASER~\citep{yang2025vlaser} analyzes the gap between embodied reasoning and policy learning, showing that out-of-domain reasoning gains may not transfer directly to control.
Knowledge Insulation~\citep{driess2025knowledge} further studies component retention during VLA training, showing that protecting the VLM representation from being degraded by action loss can benefit downstream policy learning.
Building on these studies, we further examine how a broader set of factors jointly shape the VLM representation and affect VLA initialization.

\section{Preliminaries and Study Design}
\label{sec:prelim}

This section describes the controlled study used to analyze how different VLM representations affect VLA initialization. We organize the study along three axes: capability-level embodied VQA domains and their compositions, parameter-update strategy, and robot-data pretraining. We also describe the VLA architectures, training protocol, and evaluation benchmarks.

\subsection{VLA Architectures}
\label{sec:prelim:arch}
We use two action-head designs to compare VLA initializations.
Our primary architecture follows OpenVLA-OFT~\citep{kim2025fine}: the VLM encodes the visual observation and language instruction, and a lightweight MLP decodes the hidden states into continuous action chunks.
This minimal action head makes the action policy more sensitive to differences in the VLM initialization.
We also evaluate a $\pi_0$-style variant~\citep{black2024pi_0} with a diffusion action expert to assess whether the observed patterns persist under a higher-capacity action decoder.

\subsection{Embodied VQA Domains}
\label{sec:prelim:taxonomy}

\begin{figure}[t]
  \centering
  \includegraphics[width=0.98\linewidth]{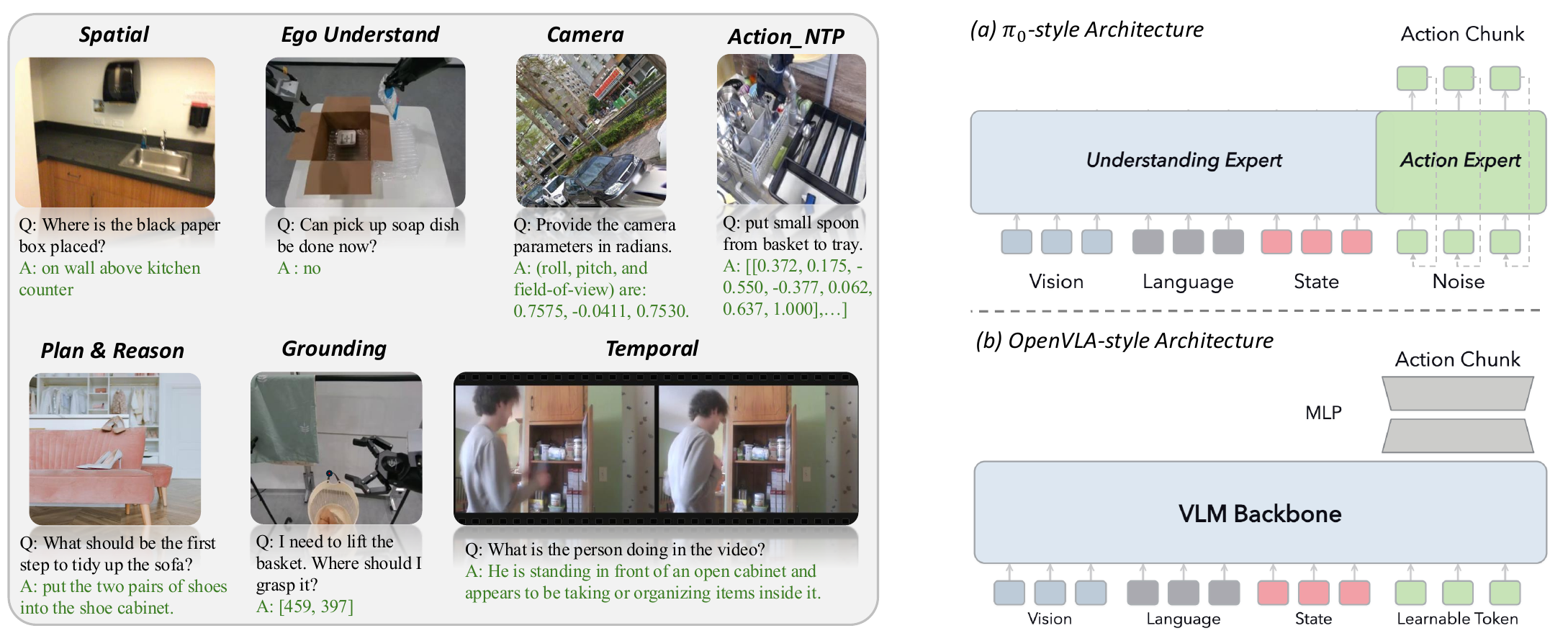}
  \vspace{-1mm}
  \caption{Illustrations of the seven embodied VQA domains and two VLA architectures.}
  \label{fig:method}
\end{figure}

We organize embodied VQA data into seven capability-oriented domains, as illustrated in Fig.~\ref{fig:method}. Specifically, \textbf{Spatial} covers relative and absolute spatial relations, orientation, and distance between entities. \textbf{Grounding} focuses on spatial referring, where the model localizes language-referred objects, actionable regions, or trajectory-relevant targets. \textbf{Plan \& Reasoning} decomposes high-level goals into subgoals and reasons about task preconditions or step ordering. \textbf{Camera Prediction} estimates camera intrinsics, extrinsics, or relative viewpoint changes from visual observations. \textbf{Egocentric Understanding} captures egocentric state information, such as hand or gripper position, held objects, and reachable objects. \textbf{Temporal Understanding} reasons over video events, including action ordering, event boundaries, and causal relations. \textbf{Action Next-Token Prediction (Action-NTP)} treats action trajectories as language-like tokens and trains the VLM to predict them autoregressively.
The data sources for each domain are listed in Appendix~\ref{app:data}.

\subsection{Two-Stage Training Pipeline}
\label{sec:prelim:setup}

We use a two-stage training pipeline to separate representation shaping from downstream action learning. In Stage~1, we adapt the base VLM on embodied VQA data to inject capability-level signals. In Stage~2, the resulting VLM initializes a VLA policy, which is then trained on action trajectories using the same downstream recipe. This separation lets us attribute differences to the Stage-1 initialization rather than to changes in action-policy training.

We also consider two update strategies for Stage~1 adaptation. \textbf{LoRA}~\citep{hu2022lora} updates only a small set of adapter parameters, preserving most original VLM weights and limiting representation drift. \textbf{Full Finetune} updates all parameters, allowing VQA supervision to reshape the representation more aggressively. Comparing the two strategies lets us study whether VLA initialization benefits more from aggressive specialization or from preserving the pretrained representation that remains useful for action learning.

\subsection{Robot-Data Pretraining}
\label{sec:prelim:pretrain}

Beyond perception-side VLM adaptation, we further study robot-data pretraining as an action-side signal for shaping VLA initialization. We use AgiBot-World-Beta~\citep{bu2025agibot_iros} as the action data source and compare robot-data-only pretraining, joint robot-data and VQA pretraining, and staged pretraining that first applies perception-side VQA adaptation and then continues with robot-data pretraining. These settings let us examine whether robot trajectories provide a useful initialization signal by themselves, whether they should be combined with perception-side VQA supervision, or whether the two signals are better injected sequentially.
We also compare LoRA and Full Finetune where applicable. This lets us assess whether action-side supervision should fully reshape the VLM backbone, or whether preserving the pretrained VLM representation remains beneficial when the injected signal comes from robot trajectories.

\subsection{Experiment Settings and Evaluation Protocol}
\label{sec:prelim:eval}

To isolate the effect of VLA initialization, we use the same observation interface, downstream training recipe, and evaluation protocol for all compared models within each benchmark and action architecture. Each policy observes a single RGB image and the language instruction; we omit proprioceptive state and action history so that differences mainly reflect the visual-language initialization. Full Stage~1/Stage~2 hyperparameters, rollout settings, and evaluation details are provided in Appendix~\ref{app:hparams}. We report the mean success rate over three evaluation seeds.

\textbf{Benchmarks.} We evaluate on three simulated benchmarks with different bottlenecks: single-arm tabletop manipulation, real-to-sim manipulation, and high-dimensional humanoid control.

\noindent
\textit{\underline{(1) Libero-10}} is the long-horizon split of Libero~\citep{liu2023libero}, containing 10 single-arm tabletop manipulation tasks. Compared with other benchmarks, it has simpler control dynamics and relatively limited scene and task diversity, with the main challenge coming from long-horizon task execution.

\noindent
\textit{\underline{(2) SimplerBridge}} is the WidowX Bridge-V2 task suite~\citep{walke2023bridgedata} in SimplerEnv~\citep{li24simpler}, a real-to-sim benchmark designed to mirror real WidowX rollouts in simulation. We evaluate on four task variants (\texttt{Pick Carrot}, \texttt{Pick Eggplant}, \texttt{Pick Spoon}, and \texttt{Stack Cube}), which introduce stronger visual and control shifts.

\noindent
\textit{\underline{(3) RoboCasa GR1 Tabletop}}~\citep{nasiriany2024robocasa} contains 24 humanoid manipulation tasks across diverse household scenes. The policy outputs a 29-dimensional action covering both arms, hands, and the waist. Its bimanual control, articulated-object interactions, and cross-scene diversity make it the most challenging benchmark in our study.

\textbf{Simulation protocol.} We use simulated benchmarks to keep the comparison reproducible and controlled. This allows us to isolate the effect of VLA initialization, whereas real-world rollouts introduce hardware, sensing, calibration, and environment variance that can obscure differences between initializations.

\section{Experiments and Analysis}
\label{sec:main}

This section examines how the three axes introduced in Sec.~\ref{sec:prelim} shape the VLM representation used for downstream VLA initialization.
We first study \textbf{capability-level embodied VQA domains and compositions}, asking which injected signals transfer positively to action learning and how different domains interact (Sec.~\ref{sec:axis1}).
We then compare \textbf{parameter-update strategies}, using LoRA and Full Finetune to analyze the trade-off between preserving the pretrained representation and specializing it toward embodied supervision (Sec.~\ref{sec:axis2}).
Finally, we study \textbf{robot-data pretraining} as an action-side adaptation signal and examine how it interacts with perception-side VQA adaptation (Sec.~\ref{sec:robot-data-pretrain}).

\subsection{Embodied VQA Capabilities for VLA Initialization}
\label{sec:axis1}

We begin with the first axis: perception-side adaptation through embodied VQA. Each VQA domain supplies a distinct capability-oriented supervision signal, so it can reshape the VLM representation in a different way before VLA training. Sec.~\ref{sec:axis1:domain} studies single-domain adaptation to identify which capabilities transfer positively to action learning; Sec.~\ref{sec:axis1:mixing} analyzes domain composition to determine when useful single-domain signals remain complementary and when they interfere with one another; Sec.~\ref{sec:axis1:quant} summarizes the main observations and practical implications.

\subsubsection{Single-Domain VQA Adaptation}
\label{sec:axis1:domain}

\paragraph{Setup.} 
To isolate the effect of the injected capability, we fix the Stage-1 update strategy to LoRA in this section. In Stage~1, we adapt Qwen3-VL-4B~\citep{bai2025qwen3} separately on each of the seven embodied VQA domains. In Stage~2, each adapted VLM backbone initializes a VLA policy, which is trained on action trajectories. We compare these initializations with two references: ``Baseline'' directly initializes the VLA from the off-the-shelf VLM checkpoint, while ``Train from scratch'' trains the VLA without pretrained VLM initialization.

\begin{table}[t]
  \caption{Success rates (\%) of single-domain VQA adaptation on three benchmarks under two action heads. Each row shows the absolute delta against its column's baseline (\textcolor{gaincolor}{green} $=$ gain, \textcolor{losscolor}{red} $=$ loss).}
  \label{tab:axis1:domain}
  \centering
  \small
  \setlength{\tabcolsep}{3.5pt}
  \begin{tabular}{@{}l cccccc@{}}
    \toprule
     & \multicolumn{3}{c}{MLP Head} & \multicolumn{3}{c}{Diffusion Expert} \\
    \cmidrule(r){2-4}\cmidrule(l){5-7}
    Domain & Libero-10 & SimplerBridge & RoboCasa & Libero-10 & SimplerBridge & RoboCasa \\
    \midrule
    \rowcolor{gray!15}
    Train from scratch & 66.6 & 19.4 & 20.3 & 68.6 & 28.9 & 30.1 \\
    Baseline           & 92.4 & \textbf{45.8} & 49.5 & 91.8 & 50.5 & 51.7 \\
    \midrule
    Spatial          & 93.0\,\gain{0.6} & 41.4\,\loss{-4.4} & 49.2\,\loss{-0.3} & 92.4\,\gain{0.6} & 49.9\,\loss{-0.6} & 50.0\,\loss{-1.7} \\
    Grounding        & 95.6\,\gain{3.2} & 44.8\,\loss{-1.0} & \textbf{50.4}\,\gain{0.9} & \textbf{94.2}\,\gain{2.4} & \textbf{50.8}\,\gain{0.3} & \textbf{52.7}\,\gain{1.0} \\
    Plan \& Reason.  & 95.2\,\gain{2.8} & 37.8\,\loss{-8.0} & 47.5\,\loss{-2.0} & 92.8\,\gain{1.0} & 48.1\,\loss{-2.4} & 51.1\,\loss{-0.6} \\
    Camera Pred.     & 93.2\,\gain{0.8} & 43.0\,\loss{-2.8} & 47.9\,\loss{-1.6} & 92.6\,\gain{0.8} & 46.6\,\loss{-3.9} & 50.7\,\loss{-1.0} \\
    Ego Unders.      & 95.3\,\gain{2.9} & 43.2\,\loss{-2.6} & 49.9\,\gain{0.4} & 93.0\,\gain{1.2} & 49.8\,\loss{-0.7} & 52.0\,\gain{0.3} \\
    Temporal         & \textbf{96.4}\,\gain{4.0} & 38.2\,\loss{-7.6} & 47.9\,\loss{-1.6} & 93.1\,\gain{1.3} & 49.1\,\loss{-1.4} & 50.7\,\loss{-1.0} \\
    Action-NTP       & 95.4\,\gain{3.0} & 44.0\,\loss{-2.0} & 50.0\,\gain{0.5} & 93.4\,\gain{1.6} & 49.6\,\loss{-0.9} & 52.3\,\gain{0.6} \\
    \bottomrule
  \end{tabular}
\end{table}

\paragraph{Results.} 
Table~\ref{tab:axis1:domain} shows three main patterns. First, pretrained VLM initialization is critical for action learning. Compared with the Baseline, training from scratch drops significantly on all benchmarks under both action heads. Second, the effect of single-domain VQA adaptation does not produce uniform gains. On Libero-10, almost every domain improves over the Baseline, with gains up to $+4.0$\% under the MLP head and $+2.4$\% under the Diffusion Expert. On SimplerBridge, the trend largely reverses: most domains fall below the Baseline, and only Grounding slightly exceeds it under the Diffusion Expert. On RoboCasa, single-domain adaptation has a more limited effect, with both gains and drops staying close to the Baseline.

Within this benchmark-dependent pattern, the injected capability also matters. Grounding is the most consistent positive case: it achieves the best performance under the Diffusion Expert across all three benchmarks, and under the MLP head it improves Libero-10 and RoboCasa while causing only a small drop on SimplerBridge. 
Egocentric Understanding and Action-NTP also provide relatively robust signals, improving in most settings and degrading less severely on SimplerBridge than several other domains. These results argue against a simple recipe in which embodied VQA adaptation uniformly improves VLA initialization. Instead, its transfer effect depends on both the downstream benchmark and the capability being injected.

\subsubsection{Multi-Domain VQA Composition}
\label{sec:axis1:mixing}

\paragraph{Setup.}
We next examine domain composition: whether data that are useful individually remain complementary when combined. Based on the single-domain results, we select Grounding, Egocentric Understanding, and Action-NTP as three relatively robust candidates. We evaluate all pairwise compositions among these domains, as well as their three-domain composition. We also add Spatial as a control domain outside this set and include the uniform seven-domain composition as a broad-coverage reference. To decouple domain composition from data scale, we fix the total data budget at 800k samples and sample evenly from the selected domains in each composition.

\begin{table}[t]
  \caption{Domain-composition success rate (\%) for three representative VQA domains.}
  \label{tab:axis1:mixing}
  \centering
  \small
  \setlength{\tabcolsep}{4pt}
  \begin{tabular}{@{}l cc cc@{}}
    \toprule
     & \multicolumn{2}{c}{MLP Head} & \multicolumn{2}{c}{Diffusion Expert} \\
    \cmidrule(lr){2-3}\cmidrule(l){4-5}
    Configuration & Libero-10 & RoboCasa & Libero-10 & RoboCasa \\
    \midrule
    Grounding                                   & \underline{95.6} & \underline{50.4} & 94.2 & \underline{52.7} \\
    Ego Unders.                                 & 95.3 & 49.9 & 93.0 & 52.0 \\
    Action-NTP                                  & 95.4 & 50.0 & 93.4 & 52.3 \\
    \midrule
    \rowcolor{gray!15}\multicolumn{5}{@{}l}{\emph{pairwise compositions}} \\
    Grounding + Ego                             & \textbf{95.7} & \textbf{51.5} & \textbf{95.8} & \textbf{53.5} \\
    Grounding + Action-NTP                      & \underline{95.2} & 50.6 & 94.5 & 51.9 \\
    Ego + Action-NTP                            & 95.0 & 50.2 & \underline{94.8} & 51.7 \\
    \midrule
    \rowcolor{gray!15}\multicolumn{5}{@{}l}{\emph{multi-domain compositions}} \\
    Grounding + Ego + Action-NTP                 & 95.0 & 49.5 & 94.1 & 51.2 \\
    Grounding + Ego + Spatial                    & 94.6 & 49.7 & 93.6 & 50.7 \\
    Grounding + Ego + Action-NTP + Spatial       & 94.5 & 48.4 & 93.0 & 50.4 \\
    Uniform 7-domain composition                 & 94.2 & 49.1 & 93.9 & 50.4 \\
    \bottomrule
  \end{tabular}
\end{table}

\paragraph{Results.}
Table~\ref{tab:axis1:mixing} shows that \{Grounding\,+\,Ego\} is the strongest composition, achieving the best performance across both action heads and benchmarks. However, this gain is specific to the domain pair. The other pairwise compositions, \{Grounding\,+\,Action-NTP\} and \{Ego\,+ \,Action-NTP\}, do not match the \{Grounding\,+\,Ego\} result and remain close to the single-domain references. Thus, combining individually useful domains is not consistently beneficial; the effect depends on which capabilities are combined.
We speculate that Grounding and Egocentric Understanding provide more compatible signals for action learning, since this compatibility does not extend monotonically as more domains are added. The three-domain composition does not improve over the best pair, adding Spatial follows a similar saturation or drop pattern, and the seven-domain composition also fails to recover the peak. These results indicate that gains from different VQA domains are not additive, and that broader embodied VQA coverage may dilute useful supervision or introduce interference.

\subsubsection{Synthesis}
\label{sec:axis1:quant}

$\bullet$~\textbf{Embodied VQA transfer depends on both the downstream bottleneck and the injected capability.}
A common intuition is that embodied VQA data should generally improve VLA initialization, but our results show that this effect is conditional.
The same adaptation produces different transfer patterns across benchmarks, suggesting that its benefit depends on the downstream bottleneck.
The injected capability also matters: Grounding, Egocentric Understanding, and Action-NTP provide more robust transfer signals, whereas other domains do not transfer consistently across settings.
Appendix~\ref{app:benchmark_analysis} further evaluates the same single-domain adaptation on Libero-10-plus
~\citep{fei2025libero} as an in-family stress test.
There, positive transfer persists, supporting the bottleneck-alignment interpretation rather than a simple benchmark-difficulty explanation.
Appendix~\ref{app:probing} provides a frozen-backbone probing analysis that is consistent with these capability-level transfer patterns.

$\bullet$~\textbf{Domain compatibility matters more than broad embodied coverage.}
Multi-domain composition does not yield additive gains from VQA supervision.
The strongest result comes from \{Grounding\,+\,Ego\}, while other pairwise and broader compositions saturate or degrade.
This suggests that useful composition depends more on compatibility between capabilities than on covering more embodied domains.
Grounding and Egocentric Understanding may provide mutually supportive signals for action learning, whereas adding additional domains can dilute useful supervision or introduce interference.

\subsection{Update Strategy for VLA Initialization}
\label{sec:axis2}

After studying which embodied VQA signals to inject, we turn to the second axis: how much the pretrained VLM representation should be updated during adaptation.
Full Finetune updates the whole backbone and can substantially reshape the original representation toward the Stage-1 embodied supervision.
In contrast, LoRA-based adaptation uses a more constrained update, limiting representation drift and preserving more of the pretrained VLM representation.
This section compares the two strategies to analyze the preservation-specialization trade-off: whether VLA initialization benefits more from aggressively specializing the VLM or from injecting embodied signals while retaining the original pretrained representation.

\begin{figure}[t]
  \centering
  \includegraphics[width=0.99\linewidth]{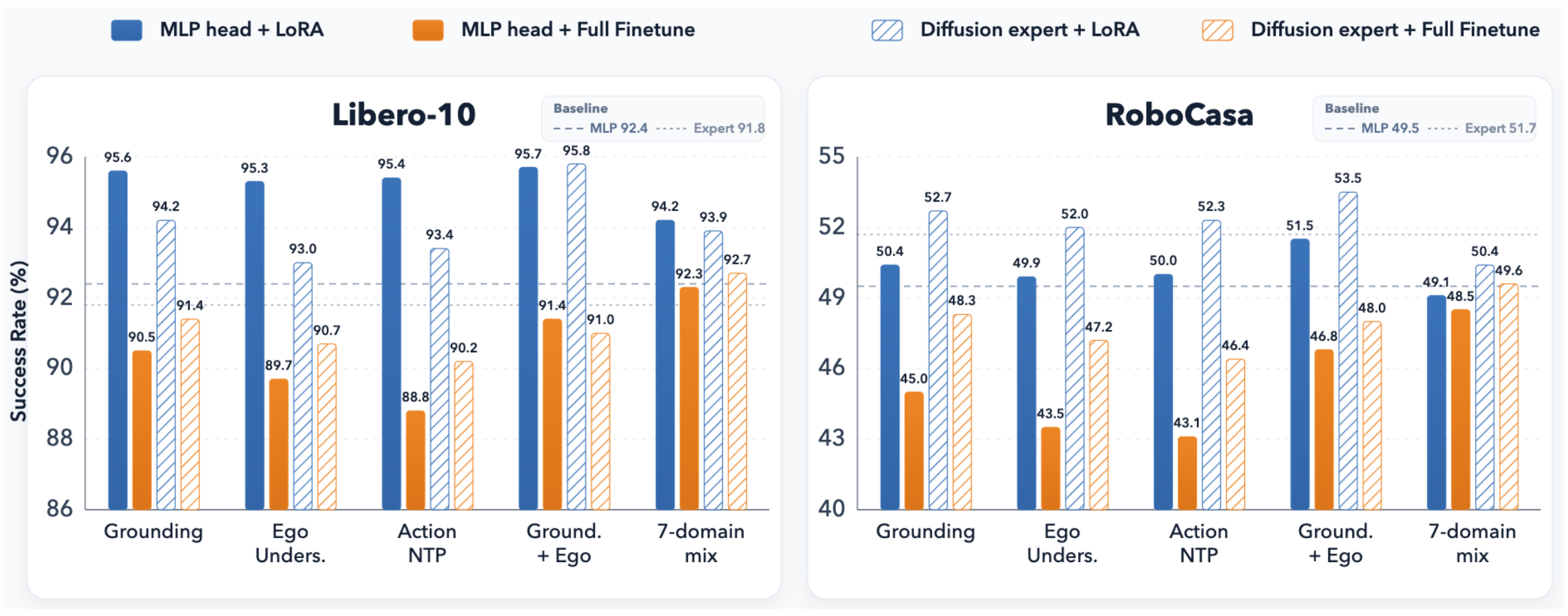}
  \caption{Effect of parameter update strategies (LoRA vs. Full Finetune) on VLA initialization.}
  \label{fig:421-exp-fig}
\end{figure}

\subsubsection{Preservation vs.\ Specialization}
\label{sec:axis2:lora}

\paragraph{Setup.}
We compare LoRA and Full Finetune while keeping the Stage-1 data and Stage-2 training recipe fixed. Specifically, both update strategies use the single-domain and multi-domain VQA settings studied in Sec.~\ref{sec:axis1} with the same training hyperparameters. For LoRA, the learned adapter weights are merged into the original VLM weights after Stage~1, and the merged checkpoint is used to initialize the VLA policy. We use Qwen3-VL-4B as the base VLM and evaluate both action heads.

\paragraph{Results.}
As shown in Fig.~\ref{fig:421-exp-fig}, LoRA consistently produces a stronger VLA initialization than Full Finetune across benchmarks and action heads. In the single-domain settings, Full Finetune falls below the Baseline, whereas LoRA improves the baseline in several cases. This suggests that full-parameter adaptation does not simply inject the target capability. It also reshapes the pretrained VLM representation in ways that can reduce its usefulness for downstream action learning.
The multi-domain Full Finetune results help separate two effects: broader supervision and representation preservation. Broader supervision performs better than single-domain Full Finetune, possibly because the broader supervision acts as a regularizer against narrow domain specialization, but it still remains below the corresponding LoRA initialization. Thus, the improvement from multi-domain Full Finetune should not be interpreted as additive domain gain. Instead, it suggests that broad full-parameter finetuning is less damaging than narrow specialization, while representation preservation remains more effective for VLA initialization.

Together, these results support a preservation-vs.-specialization interpretation. The pretrained VLM representation appears to contain transferable information that remains useful for action learning. Full Finetune can overwrite or interfere with this information, whereas LoRA preserves more of this information while injecting the embodied signal
Under this interpretation, LoRA provides a better VLA initialization because it better balances embodied specialization with representation preservation.
Appendix~\ref{app:preservation_strength} provides additional evidence from LoRA merge-strength analysis, where over-amplifying the update weakens transfer. Appendix~\ref{app:behavioral_retention} further shows that Full Finetune learns the auxiliary VQA task more strongly but loses substantially more general VLM capability and downstream VLA performance.
These diagnostics strengthen our interpretation that useful VLA initialization depends not only on learning the auxiliary embodied task, but also on retaining the pretrained VLM representation.

\subsubsection{Effect of Base VLM Strength}
\label{sec:axis2:strength}

\paragraph{Setup.}
We further examine whether the preservation effect depends on the base VLM. To separate within-family scaling from broader backbone differences, we compare Qwen3-VL-4B with Qwen3-VL-2B~\citep{bai2025qwen3} as a same-family smaller backbone, and include PaliGemma2-3B~\citep{steiner2024paligemma} as an out-of-family reference. For each base VLM, we evaluate the three single-domain settings mentioned in Sec.~\ref{sec:axis1:domain}: Grounding, Egocentric Understanding, and Action-NTP. We use the OpenVLA-OFT-style architecture and evaluate on Libero-10 and RoboCasa. We report detailed success rates as well as the average success-rate change relative to each base VLM's own baseline, averaged across the three domains.

\begin{figure*}[t]
  \centering
  \begin{minipage}[b]{0.45\textwidth}
    \vspace{0pt}
    \centering
    \includegraphics[width=\linewidth]{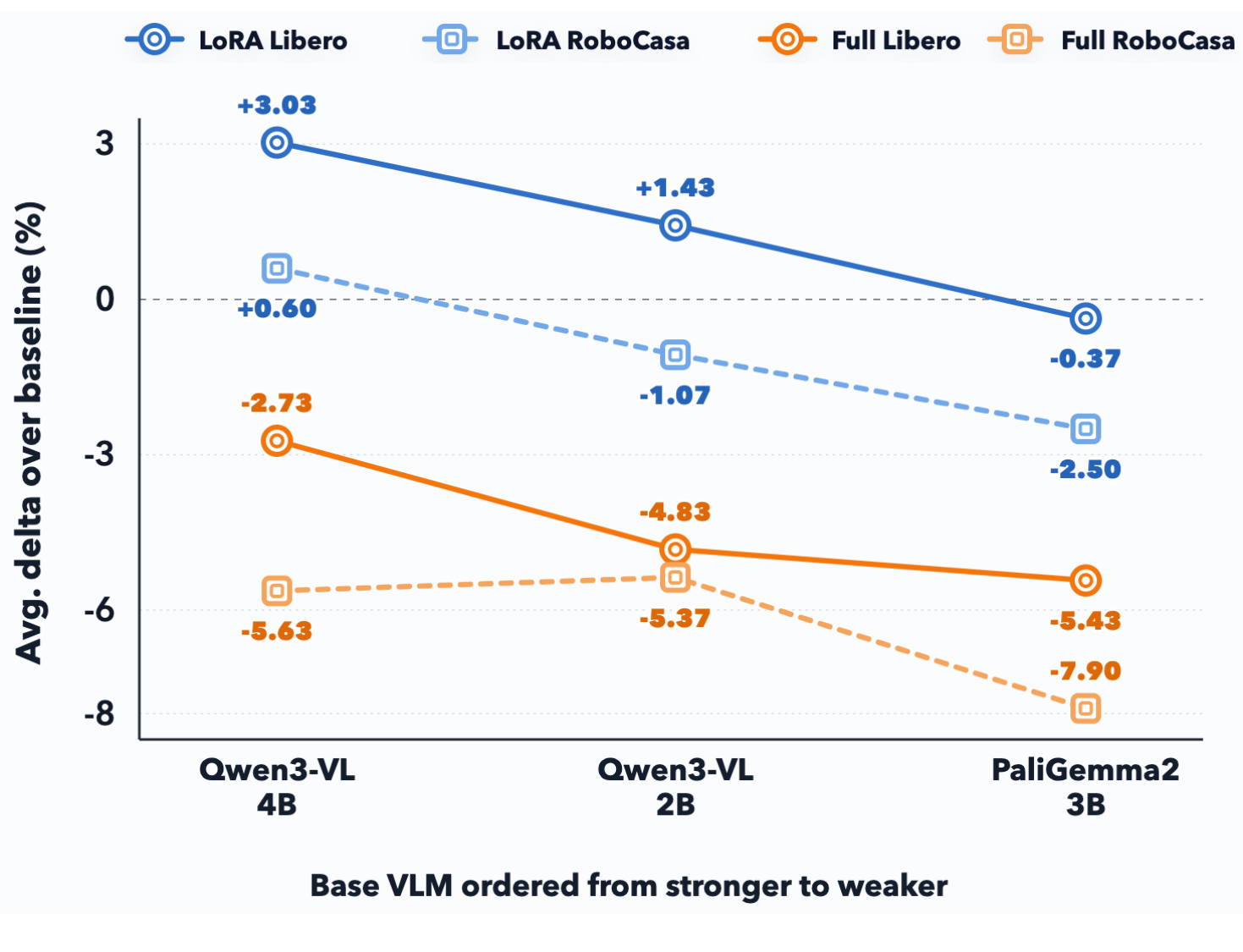}
    \par\vspace{-0.5mm}
    \centerline{\scriptsize\textbf{(a) Base-strength trend analysis}}
  \end{minipage}
  \hfill
  \begin{minipage}[b]{0.53\textwidth}
    \vspace{0pt}
    \centering
    \scriptsize
    \setlength{\tabcolsep}{1.9pt}
    \renewcommand{\arraystretch}{0.84}
    \vspace{-1.5mm}
    \begin{tabular}{@{}llcccc@{}}
      \toprule
       &  & \multicolumn{2}{c}{LoRA} & \multicolumn{2}{c}{Full Finetune} \\
      \cmidrule(r){3-4}\cmidrule(l){5-6}
      Base VLM & Domain & Libero & RoboCasa & Libero & RoboCasa \\
      \midrule
      \rowcolor{gray!15}\multirow{4}{*}{Qwen3-VL-4B} & Baseline & 92.4 & 49.5 & 92.4 & 49.5 \\
       & Grounding  & 95.6 & 50.4 & 90.5 & 45.0 \\
       & Ego Unders. & 95.3 & 49.9 & 89.7 & 43.5 \\
       & Action-NTP & 95.4 & 50.0 & 88.8 & 43.1 \\
      \midrule
      \rowcolor{gray!15}\multirow{4}{*}{Qwen3-VL-2B} & Baseline & 93.2 & 48.8 & 93.2 & 48.8 \\
       & Grounding  & 95.2 & 49.6 & 89.1 & 44.1 \\
       & Ego Unders. & 94.6 & 46.1 & 88.4 & 42.9 \\
       & Action-NTP & 94.1 & 47.5 & 87.6 & 43.3 \\
      \midrule
      \rowcolor{gray!15}\multirow{4}{*}{PaliGemma2-3B} & Baseline & 90.1 & 46.4 & 90.1 & 46.4 \\
       & Grounding  & 90.7 & 44.5 & 85.6 & 39.8 \\
       & Ego Unders. & 89.2 & 43.3 & 84.0 & 37.2 \\
       & Action-NTP & 89.3 & 43.9 & 84.4 & 38.5 \\
      \bottomrule
    \end{tabular}
    \par\vspace{4.0mm}
    \centerline{\scriptsize\textbf{(b) Detailed success rates}}
  \end{minipage}
  \vspace{-0.1mm}
  \caption{\textbf{Effect of base VLM backbones on adaptation.} (a) Average success-rate changes relative to each base VLM's own baseline. (b) Detailed success rates (\%) for the three base VLMs.}
  \label{fig:axis2:strength}
\end{figure*}

\paragraph{Results.}
Table~\ref{fig:axis2:strength}(b) shows that LoRA outperforms Full Finetune across all three base VLMs and both benchmarks, indicating that the preservation advantage is not specific to a single backbone. 
However, LoRA's gain over each model's own baseline decreases as the base VLM becomes weaker. On Libero-10, the average gain drops from $+3.03$\% for Qwen3-VL-4B to $+1.43$\% for Qwen3-VL-2B, and becomes negative for PaliGemma2-3B. RoboCasa follows the same direction, decreasing from $+0.60$\% to negative gains. This suggests that embodied capability injection through LoRA is more effective when the base VLM already provides a strong transferable representation.
Full Finetune shows a consistent negative pattern. Across all base VLMs, it falls below each model's own baseline and remains well below the corresponding LoRA initialization. This further supports the preservation view: the original pretrained representation contains transferable information useful for action learning, and overly rewriting it can weaken the resulting VLA initialization.

\subsubsection{Synthesis and Implications}
\label{sec:axis2:synthesis}

$\bullet$~\textbf{The pretrained VLM representation is important for action learning.}
Stage-1 adaptation does not simply inject embodiment-relevant capabilities; it also determines how much of the pretrained VLM representation is retained. The comparison between LoRA and Full Finetune suggests that this retained representation contains transferable information useful for downstream VLA training. LoRA provides a stronger initialization by injecting embodied signals while limiting representation drift, whereas full finetuning can over-specialize the VLM and make the resulting initialization less useful for action learning.

$\bullet$~\textbf{Base VLM strength affects adaptation gains.}
Adapter-based adaptation is more beneficial when there is a strong pretrained representation to preserve and reuse. Given the same embodied capability injection, stronger base VLMs show clearer downstream gains, while weaker backbones yield smaller or even negative gains. This suggests that LoRA's advantage also depends on the quality of the pretrained representation it preserves, rather than on the constrained update alone.

Therefore, a practical strategy is to start from a strong base VLM and use constrained adapter-based updates to inject embodiment-relevant capabilities. Full finetuning should be used cautiously unless the supervision is sufficiently broad, high-quality, and well aligned with downstream action learning.

\subsection{Robot-Data Pretraining for VLA Initialization}
\label{sec:robot-data-pretrain}

After examining perception-side VLM adaptation, we next study robot-data pretraining as an action-side route to VLA initialization, which provides direct supervision over observation-action mappings. This section examines whether robot-trajectory supervision can shape a useful VLA initialization, how it interacts with perception-side VQA adaptation, and whether representation preservation still remains important when the injected signal comes from robot data.

\paragraph{Setup.}
We use AgiBot-World-Beta~\citep{bu2025agibot_iros} as the robot-data source for Stage-1 pretraining. We compare three ways of injecting robot-data supervision: robot-only pretraining, joint robot-data and VQA pretraining, and sequential pretraining that starts from the \{Grounding\,+\,Ego\} adapted VLM in Sec.~\ref{sec:axis1:mixing}. Here, VQA denotes a mixture of general VQA data and multi-domain embodied VQA data. To identify representation preservation, we compare Full Finetune and LoRA in the relevant pretraining settings. We use Qwen3-VL-4B with the OpenVLA-OFT architecture and evaluate on RoboCasa, whose bimanual manipulation setting is closer to the AgiBot trajectories.

\begin{wraptable}{r}{0.58\linewidth}
  \vspace{-6mm}
  \caption{Effect of AgiBot robot-data pretraining. G+E denotes Grounding and Egocentric Understanding.}
  \label{tab:axis2:pretrain}
  \vspace{1mm}
  \centering
  \footnotesize
  \setlength{\tabcolsep}{2.0pt}
  \begin{tabular}{@{}>{\raggedright\arraybackslash}p{0.18\linewidth}
                  >{\raggedright\arraybackslash}p{0.34\linewidth}
                  >{\centering\arraybackslash}p{0.14\linewidth}
                  >{\centering\arraybackslash}p{0.20\linewidth}@{}}
    \toprule
    Init. VLM & Pretrain Data & Update & RoboCasa \\
    \midrule
    Base & -- & -- & 49.5 \\
    G+E & -- & -- & 51.5\,\gain{2.0} \\
    \midrule
    Base & AgiBot & Full FT & 52.0\,\gain{2.5} \\
    Base & AgiBot + VQA & Full FT & 53.2\,\gain{3.7} \\
    Base & AgiBot & LoRA r64 & 54.0\,\gain{4.5} \\
    Base & AgiBot + VQA & LoRA r64 & 52.4\,\gain{2.9} \\
    Base & AgiBot + VQA & LoRA r16 & 51.5\,\gain{2.0} \\
    \midrule
    G+E & AgiBot & LoRA r64 & \textbf{55.2}\,\gain{5.7} \\
    Base & AgiBot + G+E-VQA & LoRA r64 & 52.6\,\gain{3.1} \\
    \bottomrule
  \end{tabular}
  \vspace{-1mm}
\end{wraptable}

\paragraph{Results.}
Table~\ref{tab:axis2:pretrain} shows that all pretrained variants improve over the Base initialization.
We first compare robot-data pretraining with and without auxiliary VQA supervision under Full Finetune. Robot-data-only pretraining improves the Base initialization from $49.5$\% to $52.0$\%, while adding VQA further increases performance to $53.2$\%. This suggests that auxiliary VQA supervision can regularize full-parameter robot pretraining by helping retain perception-side VLM representations while the model learns from action trajectories.
For the LoRA-based pretraining, the results show that representation preservation still remains important even when the injected signal comes from robot data.
Robot-data-only LoRA reaches $54.0$\%, outperforming Full Finetune at $52.0$\%.
However, joint robot-VQA training under LoRA performs worse than robot-data-only LoRA. The LoRA-rank comparison supports a capacity-based explanation: when the joint robot-VQA adapter rank is reduced from 64 to 16, performance drops from $52.4$\% to $51.5$\%. This suggests that a shared LoRA adapter can face objective competition between perception-side VQA supervision and action-side supervision, especially when adapter capacity is limited.
For sequential pretraining, starting from the G+E-adapted VLM and then applying LoRA-based robot-data pretraining reaches the best performance at $55.2$\%. This indicates that a staged update pattern can help mitigate the competition between perception-side and action-side signals by injecting them sequentially. This staged pattern is promising, but its current evidence is limited to a single pretraining robot-data source, as discussed in Appendix~\ref{app:limitations}.

\section{Conclusion}
\label{sec:conclusion}

In this paper, we studied what makes a pretrained VLM representation useful for VLA initialization, framing the problem as controlled representation design before downstream action learning. Across embodied VQA domains, update strategies, and robot-data pretraining, our results show that effective initialization requires both action-relevant signal injection and preservation of the pretrained VLM representation. Embodied VQA helps only when aligned with downstream bottlenecks, full-parameter finetuning can overwrite transferable representations which is detrimental, and robot-data pretraining is stronger when combined with specific embodied VQA adaptation through staged adapters. These findings provide a nuanced understanding of the design space for VLA initialization and practical insights for constructing effective VLA initializations from pretrained VLMs.

\clearpage

\bibliographystyle{plainnat}
\bibliography{references}

@article{brohan2022rt,
  title={Rt-1: Robotics transformer for real-world control at scale},
  author={Brohan, Anthony and Brown, Noah and Carbajal, Justice and Chebotar, Yevgen and Dabis, Joseph and Finn, Chelsea and Gopalakrishnan, Keerthana and Hausman, Karol and Herzog, Alex and Hsu, Jasmine and others},
  journal={arXiv preprint arXiv:2212.06817},
  year={2022}
}

@article{driess2023palm,
  title={Palm-e: An embodied multimodal language model},
  author={Driess, Danny and Xia, Fei and Sajjadi, Mehdi SM and Lynch, Corey and Chowdhery, Aakanksha and Ichter, Brian and Wahid, Ayzaan and Tompson, Jonathan and Vuong, Quan and Yu, Tianhe and others},
  journal={arXiv preprint arXiv:2303.03378},
  year={2023}
}

@article{huang2022inner,
  title={Inner monologue: Embodied reasoning through planning with language models},
  author={Huang, Wenlong and Xia, Fei and Xiao, Ted and Chan, Harris and Liang, Jacky and Florence, Pete and Zeng, Andy and Tompson, Jonathan and Mordatch, Igor and Chebotar, Yevgen and others},
  journal={arXiv preprint arXiv:2207.05608},
  year={2022}
}

@article{brohan2023rt,
  title={Rt-2: Vision-language-action models transfer web knowledge to robotic control},
  author={Brohan, Anthony and Brown, Noah and Carbajal, Justice and Chebotar, Yevgen and Chen, Xi and Choromanski, Krzysztof and Ding, Tianli and Driess, Danny and Dubey, Avinava and Finn, Chelsea and others},
  journal={arXiv preprint arXiv:2307.15818},
  year={2023}
}

@article{mu2024embodiedgpt,
  title={Embodiedgpt: Vision-language pre-training via embodied chain of thought},
  author={Mu, Yao and Zhang, Qinglong and Hu, Mengkang and Wang, Wenhai and Ding, Mingyu and Jin, Jun and Wang, Bin and Dai, Jifeng and Qiao, Yu and Luo, Ping},
  journal={Advances in Neural Information Processing Systems},
  volume={36},
  year={2024}
}

@article{kim24openvla,
    title={OpenVLA: An Open-Source Vision-Language-Action Model},
    author={{Moo Jin} Kim and Karl Pertsch and Siddharth Karamcheti and Ted Xiao and Ashwin Balakrishna and Suraj Nair and Rafael Rafailov and Ethan Foster and Grace Lam and Pannag Sanketi and Quan Vuong and Thomas Kollar and Benjamin Burchfiel and Russ Tedrake and Dorsa Sadigh and Sergey Levine and Percy Liang and Chelsea Finn},
    journal = {arXiv preprint arXiv:2406.09246},
    year={2024},
}

@article{o2023open,
  title={Open x-embodiment: Robotic learning datasets and rt-x models},
  author={O'Neill, Abby and Rehman, Abdul and Gupta, Abhinav and Maddukuri, Abhiram and Gupta, Abhishek and Padalkar, Abhishek and Lee, Abraham and Pooley, Acorn and Gupta, Agrim and Mandlekar, Ajay and others},
  journal={arXiv preprint arXiv:2310.08864},
  year={2023}
}

@article{black2024pi_0,
  title={$pi\_0 $: A Vision-Language-Action Flow Model for General Robot Control},
  author={Black, Kevin and Brown, Noah and Driess, Danny and Esmail, Adnan and Equi, Michael and Finn, Chelsea and Fusai, Niccolo and Groom, Lachy and Hausman, Karol and Ichter, Brian and others},
  journal={arXiv preprint arXiv:2410.24164},
  year={2024}
}

@inproceedings{walke2023bridgedata,
    title={BridgeData V2: A Dataset for Robot Learning at Scale},
    author={Walke, Homer and Black, Kevin and Lee, Abraham and Kim, Moo Jin and Du, Max and Zheng, Chongyi and Zhao, Tony and Hansen-Estruch, Philippe and Vuong, Quan and He, Andre and Myers, Vivek and Fang, Kuan and Finn, Chelsea and Levine, Sergey},
    booktitle={Conference on Robot Learning (CoRL)},
    year={2023}
}

@inproceedings{chen2024spatialvlm,
  title={Spatialvlm: Endowing vision-language models with spatial reasoning capabilities},
  author={Chen, Boyuan and Xu, Zhuo and Kirmani, Sean and Ichter, Brain and Sadigh, Dorsa and Guibas, Leonidas and Xia, Fei},
  booktitle={Proceedings of the IEEE/CVF Conference on Computer Vision and Pattern Recognition},
  pages={14455--14465},
  year={2024}
}

@article{liu2023libero,
  title={LIBERO: Benchmarking Knowledge Transfer for Lifelong Robot Learning},
  author={Liu, Bo and Zhu, Yifeng and Gao, Chongkai and Feng, Yihao and Liu, Qiang and Zhu, Yuke and Stone, Peter},
  journal={arXiv preprint arXiv:2306.03310},
  year={2023}
}

@article{li24simpler,
         title={Evaluating Real-World Robot Manipulation Policies in Simulation},
         author={Xuanlin Li and Kyle Hsu and Jiayuan Gu and Karl Pertsch and Oier Mees and Homer Rich Walke and Chuyuan Fu and Ishikaa Lunawat and Isabel Sieh and Sean Kirmani and Sergey Levine and Jiajun Wu and Chelsea Finn and Hao Su and Quan Vuong and Ted Xiao},
         journal = {arXiv preprint arXiv:2405.05941},
         year={2024}
}

@article{RoboBrain2.0TechnicalReport,
    title={RoboBrain 2.0 Technical Report},
    author={BAAI RoboBrain Team},
    journal={arXiv preprint arXiv:2507.02029},
    year={2025}
}

@article{steiner2024paligemma,
  title={Paligemma 2: A family of versatile vlms for transfer},
  author={Steiner, Andreas and Pinto, Andr{\'e} Susano and Tschannen, Michael and Keysers, Daniel and Wang, Xiao and Bitton, Yonatan and Gritsenko, Alexey and Minderer, Matthias and Sherbondy, Anthony and Long, Shangbang and others},
  journal={arXiv preprint arXiv:2412.03555},
  year={2024}
}

@article{yuan2024robopoint,
  title={Robopoint: A vision-language model for spatial affordance prediction for robotics},
  author={Yuan, Wentao and Duan, Jiafei and Blukis, Valts and Pumacay, Wilbert and Krishna, Ranjay and Murali, Adithyavairavan and Mousavian, Arsalan and Fox, Dieter},
  journal={arXiv preprint arXiv:2406.10721},
  year={2024}
}

@article{feng2025towards,
  title={Towards Visuospatial Cognition via Hierarchical Fusion of Visual Experts},
  author={Feng, Qi},
  journal={arXiv preprint arXiv:2505.12363},
  year={2025}
}

@misc{chen2025robo2vlmvisualquestionanswering,
      title={Robo2VLM: Visual Question Answering from Large-Scale In-the-Wild Robot Manipulation Datasets}, 
      author={Kaiyuan Chen and Shuangyu Xie and Zehan Ma and Ken Goldberg},
      year={2025},
      eprint={2505.15517},
      archivePrefix={arXiv},
      primaryClass={cs.RO},
      url={https://arxiv.org/abs/2505.15517}, 
}

@article{pertsch2025fast,
  title={Fast: Efficient action tokenization for vision-language-action models},
  author={Pertsch, Karl and Stachowicz, Kyle and Ichter, Brian and Driess, Danny and Nair, Suraj and Vuong, Quan and Mees, Oier and Finn, Chelsea and Levine, Sergey},
  journal={arXiv preprint arXiv:2501.09747},
  year={2025}
}

@article{wen2025dexvla,
  title={Dexvla: Vision-language model with plug-in diffusion expert for general robot control},
  author={Wen, Junjie and Zhu, Yichen and Li, Jinming and Tang, Zhibin and Shen, Chaomin and Feng, Feifei},
  journal={arXiv preprint arXiv:2502.05855},
  year={2025}
}

@article{cui2025openhelix,
  title={Openhelix: A short survey, empirical analysis, and open-source dual-system vla model for robotic manipulation},
  author={Cui, Can and Ding, Pengxiang and Song, Wenxuan and Bai, Shuanghao and Tong, Xinyang and Ge, Zirui and Suo, Runze and Zhou, Wanqi and Liu, Yang and Jia, Bofang and others},
  journal={arXiv preprint arXiv:2505.03912},
  year={2025}
}

@article{intelligence2025pi_,
  title={pi0.5: a Vision-Language-Action Model with Open-World Generalization},
  author={Intelligence, Physical and Black, Kevin and Brown, Noah and Darpinian, James and Dhabalia, Karan and Driess, Danny and Esmail, Adnan and Equi, Michael and Finn, Chelsea and Fusai, Niccolo and others},
  journal={arXiv preprint arXiv:2504.16054},
  year={2025}
}

@inproceedings{sermanet2024robovqa,
  title={Robovqa: Multimodal long-horizon reasoning for robotics},
  author={Sermanet, Pierre and Ding, Tianli and Zhao, Jeffrey and Xia, Fei and Dwibedi, Debidatta and Gopalakrishnan, Keerthana and Chan, Christine and Dulac-Arnold, Gabriel and Maddineni, Sharath and Joshi, Nikhil J and others},
  booktitle={2024 IEEE International Conference on Robotics and Automation (ICRA)},
  pages={645--652},
  year={2024},
  organization={IEEE}
}

@article{ouyang2025spacer,
  title={SpaceR: Reinforcing MLLMs in Video Spatial Reasoning},
  author={Ouyang, Kun and Liu, Yuanxin and Wu, Haoning and Liu, Yi and Zhou, Hao and Zhou, Jie and Meng, Fandong and Sun, Xu},
  journal={arXiv preprint arXiv:2504.01805},
  year={2025}
}

@article{zhou2025roborefer,
  title={RoboRefer: Towards Spatial Referring with Reasoning in Vision-Language Models for Robotics},
  author={Zhou, Enshen and An, Jingkun and Chi, Cheng and Han, Yi and Rong, Shanyu and Zhang, Chi and Wang, Pengwei and Wang, Zhongyuan and Huang, Tiejun and Sheng, Lu and others},
  journal={arXiv preprint arXiv:2506.04308},
  year={2025}
}

@article{bai2025qwen3,
  title={Qwen3-VL Technical Report},
  author={Bai, Shuai and Cai, Yuxuan and Chen, Ruizhe and Chen, Keqin and Chen, Xionghui and Cheng, Zesen and Deng, Lianghao and Ding, Wei and Gao, Chang and Ge, Chunjiang and others},
  journal={arXiv preprint arXiv:2511.21631},
  year={2025}
}

@article{driess2025knowledge,
  title={Knowledge insulating vision-language-action models: Train fast, run fast, generalize better},
  author={Driess, Danny and Springenberg, Jost Tobias and Ichter, Brian and Yu, Lili and Li-Bell, Adrian and Pertsch, Karl and Ren, Allen Z and Walke, Homer and Vuong, Quan and Shi, Lucy Xiaoyang and others},
  journal={arXiv preprint arXiv:2505.23705},
  year={2025}
}

@article{kim2025fine,
  title={Fine-tuning vision-language-action models: Optimizing speed and success},
  author={Kim, Moo Jin and Finn, Chelsea and Liang, Percy},
  journal={arXiv preprint arXiv:2502.19645},
  year={2025}
}

@article{bjorck2025gr00t,
  title={Gr00t n1: An open foundation model for generalist humanoid robots},
  author={Bjorck, Johan and Casta{\~n}eda, Fernando and Cherniadev, Nikita and Da, Xingye and Ding, Runyu and Fan, Linxi and Fang, Yu and Fox, Dieter and Hu, Fengyuan and Huang, Spencer and others},
  journal={arXiv preprint arXiv:2503.14734},
  year={2025}
}

@article{zheng2025x,
  title={X-vla: Soft-prompted transformer as scalable cross-embodiment vision-language-action model},
  author={Zheng, Jinliang and Li, Jianxiong and Wang, Zhihao and Liu, Dongxiu and Kang, Xirui and Feng, Yuchun and Zheng, Yinan and Zou, Jiayin and Chen, Yilun and Zeng, Jia and others},
  journal={arXiv preprint arXiv:2510.10274},
  year={2025}
}

@article{zhang2026vlm4vla,
  title={VLM4VLA: Revisiting Vision-Language-Models in Vision-Language-Action Models},
  author={Zhang, Jianke and Chen, Xiaoyu and Wang, Qiuyue and Li, Mingsheng and Guo, Yanjiang and Hu, Yucheng and Zhang, Jiajun and Bai, Shuai and Lin, Junyang and Chen, Jianyu},
  journal={arXiv preprint arXiv:2601.03309},
  year={2026}
}

@article{yang2025vlaser,
  title={Vlaser: Vision-Language-Action Model with Synergistic Embodied Reasoning},
  author={Yang, Ganlin and Zhang, Tianyi and Hao, Haoran and Wang, Weiyun and Liu, Yibin and Wang, Dehui and Chen, Guanzhou and Cai, Zijian and Chen, Junting and Su, Weijie and others},
  journal={arXiv preprint arXiv:2510.11027},
  year={2025}
}

@inproceedings{bu2025agibot_iros,
  title={Agibot world colosseo: A large-scale manipulation platform for scalable and intelligent embodied systems},
  author={Bu, Qingwen and Cai, Jisong and Chen, Li and Cui, Xiuqi and Ding, Yan and Feng, Siyuan and He, Xindong and Huang, Xu and others},
  booktitle={2025 IEEE/RSJ International Conference on Intelligent Robots and Systems (IROS)},
  year={2025},
  organization={IEEE}
}

@article{nasiriany2024robocasa,
  title={Robocasa: Large-scale simulation of everyday tasks for generalist robots},
  author={Nasiriany, Soroush and Maddukuri, Abhiram and Zhang, Lance and Parikh, Adeet and Lo, Aaron and Joshi, Abhishek and Mandlekar, Ajay and Zhu, Yuke},
  journal={arXiv preprint arXiv:2406.02523},
  year={2024}
}

@article{geminirobotics,
  title={Gemini robotics: Bringing ai into the physical world},
  author={Team, Gemini Robotics and Abeyruwan, Saminda and Ainslie, Joshua and Alayrac, Jean-Baptiste and Arenas, Montserrat Gonzalez and Armstrong, Travis and Balakrishna, Ashwin and Baruch, Robert and Bauza, Maria and Blokzijl, Michiel and others},
  journal={arXiv preprint arXiv:2503.20020},
  year={2025}
}

@misc{cosmosr1,
    title = {Cosmos-Reason1: From Physical Common Sense To Embodied Reasoning},
    author = {NVIDIA and Azzolini, Alisson and Brandon, Hannah and Chattopadhyay, Prithvijit and Chen, Huayu and Chu, Jinju and Cui, Yin and Diamond, Jenna and Ding, Yifan and Ferroni, Francesco and Govindaraju, Rama and Gu, Jinwei and Gururani, Siddharth and El Hanafi, Imad and Hao, Zekun and Huffman, Jacob and Jin, Jingyi and Johnson, Brendan and Khan, Rizwan and Kurian, George and Lantz, Elena and Lee, Nayeon and Li, Zhaoshuo and Li, Xuan and Lin, Tsung-Yi and Lin, Yen-Chen and Liu, Ming-Yu and Mathau, Andrew and Ni, Yun and Pavao, Lindsey and Ping, Wei and Romero, David W. and Smelyanskiy, Misha and Song, Shuran and Tchapmi, Lyne and Wang, Andrew Z. and Wang, Boxin and Wang, Haoxiang and Wei, Fangyin and Xu, Jiashu and Xu, Yao and Yang, Xiaodong and Yang, Zhuolin and Zeng, Xiaohui and Zhang, Zhe},
    journal = {arXiv preprint arXiv:2503.15558},
    year = {2025},
    url = {https://arxiv.org/abs/2503.15558}
  }

@article{vlm3R,
  title={VLM-3R: Vision-Language Models Augmented with Instruction-Aligned 3D Reconstruction},
  author={Fan, Zhiwen and Zhang, Jian and Li, Renjie and Zhang, Junge and Chen, Runjin and Hu, Hezhen and Wang, Kevin and Qu, Huaizhi and Wang, Dilin and Yan, Zhicheng and others},
  journal={arXiv preprint arXiv:2505.20279},
  year={2025}
}

@article{hu2022lora,
  title={Lora: Low-rank adaptation of large language models.},
  author={Hu, Edward J and Shen, Yelong and Wallis, Phillip and Allen-Zhu, Zeyuan and Li, Yuanzhi and Wang, Shean and Wang, Liang and Chen, Weizhu and others},
  journal={Iclr},
  volume={1},
  number={2},
  pages={3},
  year={2022}
}

@misc{VQASynth,
  author = {remyxai},
  title = {VQASynth},
  year = {2024},
  note = {GitHub repository},
  url = {https://github.com/remyxai/VQASynth/tree/main}
}

@article{batra2025spatialthinker,
  title = {SpatialThinker: Reinforcing 3D Reasoning in Multimodal LLMs via
Spatial Rewards},
  author = {Batra, Hunar and Tu, Haoqin and Chen, Hardy and Lin, Yuanze and
Xie, Cihang and Clark, Ronald},
  journal = {arXiv preprint arXiv:2511.07403},
  year = {2025},
  url = {https://arxiv.org/abs/2511.07403}
}

@article{cheng2024spatialrgpt,
  title={Spatialrgpt: Grounded spatial reasoning in vision-language models},
  author={Cheng, An-Chieh and Yin, Hongxu and Fu, Yang and Guo, Qiushan and Yang, Ruihan and Kautz, Jan and Wang, Xiaolong and Liu, Sifei},
  journal={Advances in Neural Information Processing Systems},
  volume={37},
  pages={135062--135093},
  year={2024}
}

@article{cai2025scaling,
  title={Scaling spatial intelligence with multimodal foundation models},
  author={Cai, Zhongang and Wang, Ruisi and Gu, Chenyang and Pu, Fanyi and Xu, Junxiang and Wang, Yubo and Yin, Wanqi and Yang, Zhitao and Wei, Chen and Sun, Qingping and others},
  journal={arXiv preprint arXiv:2511.13719},
  year={2025}
}

@inproceedings{cai2025spatialbot,
  title={Spatialbot: Precise spatial understanding with vision language models},
  author={Cai, Wenxiao and Ponomarenko, Iaroslav and Yuan, Jianhao and Li, Xiaoqi and Yang, Wankou and Dong, Hao and Zhao, Bo},
  booktitle={2025 IEEE International Conference on Robotics and Automation (ICRA)},
  pages={9490--9498},
  year={2025},
  organization={IEEE}
}

@article{yang2025visual,
  title={Visual spatial tuning},
  author={Yang, Rui and Zhu, Ziyu and Li, Yanwei and Huang, Jingjia and Yan, Shen and Zhou, Siyuan and Liu, Zhe and Li, Xiangtai and Li, Shuangye and Wang, Wenqian and others},
  journal={arXiv preprint arXiv:2511.05491},
  year={2025}
}

@article{yang2025cambrian,
  title={Cambrian-S: Towards Spatial Supersensing in Video},
  author={Yang, Shusheng and Yang, Jihan and Huang, Pinzhi and Brown, Ellis and Yang, Zihao and Yu, Yue and Tong, Shengbang and Zheng, Zihan and Xu, Yifan and Wang, Muhan and Lu, Danhao and Fergus, Rob and LeCun, Yann and Fei-Fei, Li and Xie, Saining},
  journal={arXiv preprint arXiv:2511.04670},
  year={2025}
}

@article{liu2025spatial,
  title={Spatial-ssrl: Enhancing spatial understanding via self-supervised reinforcement learning},
  author={Liu, Yuhong and Zhang, Beichen and Zang, Yuhang and Cao, Yuhang and Xing, Long and Dong, Xiaoyi and Duan, Haodong and Lin, Dahua and Wang, Jiaqi},
  journal={arXiv preprint arXiv:2510.27606},
  year={2025}
}

@inproceedings{azuma_2022_CVPR,
  title={ScanQA: 3D Question Answering for Spatial Scene Understanding},
  author={Azuma, Daichi and Miyanishi, Taiki and Kurita, Shuhei and Kawanabe, Motoaki},
  booktitle={Proceedings of the IEEE/CVF Conference on Computer Vision and Pattern Recognition (CVPR)},
  year={2022}
}

@inproceedings{song2025robospatial,
  title={Robospatial: Teaching spatial understanding to 2d and 3d vision-language models for robotics},
  author={Song, Chan Hee and Blukis, Valts and Tremblay, Jonathan and Tyree, Stephen and Su, Yu and Birchfield, Stan},
  booktitle={Proceedings of the Computer Vision and Pattern Recognition Conference},
  pages={15768--15780},
  year={2025}
}

@inproceedings{deitke2025molmo,
  title={Molmo and pixmo: Open weights and open data for state-of-the-art vision-language models},
  author={Deitke, Matt and Clark, Christopher and Lee, Sangho and Tripathi, Rohun and Yang, Yue and Park, Jae Sung and Salehi, Mohammadreza and Muennighoff, Niklas and Lo, Kyle and Soldaini, Luca and others},
  booktitle={Proceedings of the Computer Vision and Pattern Recognition Conference},
  pages={91--104},
  year={2025}
}

@inproceedings{tang2025roboafford,
  title={Roboafford: A dataset and benchmark for enhancing object and spatial affordance learning in robot manipulation},
  author={Tang, Yingbo and Zhang, Lingfeng and Zhang, Shuyi and Zhao, Yinuo and Hao, Xiaoshuai},
  booktitle={Proceedings of the 33rd ACM International Conference on Multimedia},
  pages={12706--12713},
  year={2025}
}

@article{qu2025eo,
  title={Eo-1: Interleaved vision-text-action pretraining for general robot control},
  author={Qu, Delin and Song, Haoming and Chen, Qizhi and Chen, Zhaoqing and Gao, Xianqiang and Ye, Xinyi and Lv, Qi and Shi, Modi and Ren, Guanghui and Ruan, Cheng and others},
  journal={arXiv preprint arXiv:2508.21112},
  year={2025}
}

@inproceedings{lu2023vl,
  title={Vl-grasp: a 6-dof interactive grasp policy for language-oriented objects in cluttered indoor scenes},
  author={Lu, Yuhao and Fan, Yixuan and Deng, Beixing and Liu, Fangfu and Li, Yali and Wang, Shengjin},
  booktitle={2023 IEEE/RSJ International Conference on Intelligent Robots and Systems (IROS)},
  pages={976--983},
  year={2023},
  organization={IEEE}
}

@inproceedings{ji2025robobrain,
  title={Robobrain: A unified brain model for robotic manipulation from abstract to concrete},
  author={Ji, Yuheng and Tan, Huajie and Shi, Jiayu and Hao, Xiaoshuai and Zhang, Yuan and Zhang, Hengyuan and Wang, Pengwei and Zhao, Mengdi and Mu, Yao and An, Pengju and others},
  booktitle={Proceedings of the IEEE/CVF Conference on Computer Vision and Pattern Recognition},
  pages={1724--1734},
  year={2025}
}

@article{liao2025thinking,
  title={Thinking with Camera: A Unified Multimodal Model for Camera-Centric Understanding and Generation},
  author={Liao, Kang and Wu, Size and Wu, Zhonghua and Jin, Linyi and Wang, Chao and Wang, Yikai and Wang, Fei and Li, Wei and Loy, Chen Change},
  journal={arXiv preprint arXiv:2510.08673},
  year={2025}
}

@article{pei2025egothinker,
  title={Egothinker: Unveiling egocentric reasoning with spatio-temporal cot},
  author={Pei, Baoqi and Huang, Yifei and Xu, Jilan and He, Yuping and Chen, Guo and Wu, Fei and Qiao, Yu and Pang, Jiangmiao},
  journal={arXiv preprint arXiv:2510.23569},
  year={2025}
}

@article{jia2022egotaskqa,
  title={Egotaskqa: Understanding human tasks in egocentric videos},
  author={Jia, Baoxiong and Lei, Ting and Zhu, Song-Chun and Huang, Siyuan},
  journal={Advances in Neural Information Processing Systems},
  volume={35},
  pages={3343--3360},
  year={2022}
}

@misc{feng2025vica2,
      title={Towards Visuospatial Cognition via Hierarchical Fusion of Visual Experts}, 
      author={Feng, Qi},
      publisher={arXiv:2505.12363},
      year={2025},
}

@inproceedings{o2024open,
  title={Open x-embodiment: Robotic learning datasets and rt-x models: Open x-embodiment collaboration 0},
  author={O’Neill, Abby and Rehman, Abdul and Maddukuri, Abhiram and Gupta, Abhishek and Padalkar, Abhishek and Lee, Abraham and Pooley, Acorn and Gupta, Agrim and Mandlekar, Ajay and Jain, Ajinkya and others},
  booktitle={2024 IEEE International Conference on Robotics and Automation (ICRA)},
  pages={6892--6903},
  year={2024},
  organization={IEEE}
}

@article{fei2025libero,
  title={Libero-plus: In-depth robustness analysis of vision-language-action models},
  author={Fei, Senyu and Wang, Siyin and Shi, Junhao and Dai, Zihao and Cai, Jikun and Qian, Pengfang and Ji, Li and He, Xinzhe and Zhang, Shiduo and Fei, Zhaoye and others},
  journal={arXiv preprint arXiv:2510.13626},
  year={2025}
}

@article{community2026starvla,
  title={StarVLA: A Lego-like Codebase for Vision-Language-Action Model Developing},
  author={Community, StarVLA},
  journal={arXiv preprint arXiv:2604.05014},
  year={2026},
  url={https://github.com/starVLA/starVLA}
}

@article{singh2025openai,
  title={Openai gpt-5 system card},
  author={Singh, Aaditya and Fry, Adam and Perelman, Adam and Tart, Adam and Ganesh, Adi and El-Kishky, Ahmed and McLaughlin, Aidan and Low, Aiden and Ostrow, AJ and Ananthram, Akhila and others},
  journal={arXiv preprint arXiv:2601.03267},
  year={2025}
}

@inproceedings{liu2024mmbench,
  title={Mmbench: Is your multi-modal model an all-around player?},
  author={Liu, Yuan and Duan, Haodong and Zhang, Yuanhan and Li, Bo and Zhang, Songyang and Zhao, Wangbo and Yuan, Yike and Wang, Jiaqi and He, Conghui and Liu, Ziwei and others},
  booktitle={European conference on computer vision},
  pages={216--233},
  year={2024},
  organization={Springer}
}

@article{chen2024we,
  title={Are We on the Right Way for Evaluating Large Vision-Language Models?},
  author={Chen, Lin and Li, Jinsong and Dong, Xiaoyi and Zhang, Pan and Zang, Yuhang and Chen, Zehui and Duan, Haodong and Wang, Jiaqi and Qiao, Yu and Lin, Dahua and others},
  journal={arXiv preprint arXiv:2403.20330},
  year={2024}
}

\clearpage
\appendix

\section*{Appendix}
\addcontentsline{toc}{section}{Appendix}

\section{Implementation Detail}

\subsection{VQA Data Sources}
\label{app:data}

This section summarizes the embodied VQA data used for Stage-1 VLM adaptation.
We group the data into seven capability-oriented domains and list the source datasets in Table~\ref{tab:app:data:sources}.
The reported counts denote the available candidate pool after preprocessing.
In Sec.~\ref{sec:axis1:domain}, each single-domain model is trained with 800K samples, sampled uniformly across the source datasets within that domain.
For multi-domain composition in Sec.~\ref{sec:axis1:mixing}, we keep a fixed total budget and sample evenly across the selected domains.

\begin{table*}[h]
  \centering
  \footnotesize
  \setlength{\tabcolsep}{5.5pt}
  \renewcommand{\arraystretch}{1.12}
  \caption{Stage-1 embodied VQA data pool by capability domain.}
  \label{tab:app:data:sources}
  \begin{tabular}{@{}>{\raggedright\arraybackslash}p{0.15\textwidth}
                  >{\raggedright\arraybackslash}p{0.66\textwidth}
                  >{\raggedleft\arraybackslash}p{0.13\textwidth}@{}}
    \toprule
    \textbf{Domain} & \textbf{Main sources} & \textbf{Samples} \\
    \midrule
    Spatial
    & SpaceLLaVA~\citep{VQASynth}, SpaceThinker~\citep{VQASynth}, OpenSpaces~\citep{VQASynth}, SpatialThinker~\citep{batra2025spatialthinker}, SpatialRGPT~\citep{cheng2024spatialrgpt}, SenseNova-SI~\citep{cai2025scaling}, SpatialQA~\citep{cai2025spatialbot}, VST-500K~\citep{yang2025visual}, VSI-590K~\citep{yang2025cambrian}, Spatial-SSRL~\citep{liu2025spatial}, ScanQA~\citep{azuma_2022_CVPR}, SpaceR~\citep{ouyang2025spacer}, RoboSpatial~\citep{song2025robospatial}
    & 5{,}688K \\
    \midrule
    Grounding
    & PixMo Points~\citep{deitke2025molmo}, RoboPoint~\citep{yuan2024robopoint}, RoboRefer~\citep{zhou2025roborefer}, RoboAfford~\citep{tang2025roboafford}, EO-Data1.5M-ground~\citep{qu2025eo}, RoboRefIt~\citep{lu2023vl}, ShareRobot-affordance~\citep{ji2025robobrain}
    & 4{,}438K \\
    \midrule
    Plan \& Reasoning
    & RoboRefer reasoning~\citep{zhou2025roborefer}, and VLM-3R planning~\citep{vlm3R}
    & 815K \\
    \midrule
    Camera Prediction
    & Puffin-4M~\citep{liao2025thinking}, VSI-590k-Camera~\citep{yang2025cambrian}
    & 8{,}301K \\
    \midrule
    Ego Unders.
    & Robo2VLM~\citep{chen2025robo2vlmvisualquestionanswering}, EgoThinker~\citep{pei2025egothinker}, EgoTaskQA~\citep{jia2022egotaskqa}, EO-Data1.5M~\citep{qu2025eo}, ShareRobot~\citep{ji2025robobrain}
    & 4{,}597K \\
    \midrule
    Temporal
    & VSI-590K video~\citep{yang2025cambrian}, VICA-332K~\citep{feng2025vica2}, VLM-3R-video~\citep{vlm3R}, SpaceR~\citep{ouyang2025spacer}
    & 998K \\
    \midrule
    Action-NTP
    & OpenX-Embodiment~\citep{o2024open}, AgiBot-World-Beta~\citep{bu2025agibot_iros}
    & 1200K \\
    \bottomrule
  \end{tabular}
  \renewcommand{\arraystretch}{1.0}
\end{table*}

\subsection{Training and Evaluation Details}
\label{app:hparams}

This section reports the implementation details for Stage~1 VLM adaptation, Stage~2 VLA training, and robot-data pretraining.
Across each controlled comparison, we keep all non-targeted settings fixed within the same benchmark and action architecture, so that performance differences primarily reflect the initialization factor under study.

\noindent\textbf{Stage-1 VLM adaptation.}
For embodied VQA adaptation, each model is trained for one epoch over the sampled Stage-1 VQA data.
We use AdamW with learning rate $5{\times}10^{-5}$, global batch size 128, weight decay 0, warmup ratio 0.03, and a cosine learning-rate schedule.
In the LoRA-vs.-Full Finetune comparisons, both update strategies use the same data budget, optimizer, learning rate, batch size, and number of epochs.
For LoRA, we use rank $r=16$ and scaling factor $\alpha=32$.
LoRA adapters are inserted into every linear layer of the LLM, and the final one-quarter of the vision encoder layers are additionally unfrozen; all remaining backbone parameters are kept frozen.

\noindent\textbf{Stage-2 VLA training.}
Each Stage-1 checkpoint is used to initialize a action policy, which is then trained on specific action trajectories.
We build the VLA model and training pipeline mainly based on the StarVLA codebase framework~\citep{community2026starvla}.
Table~\ref{tab:app:stage2:hparams} summarizes the Stage~2 training and evaluation settings for Libero-10, SimplerBridge, and RoboCasa.
Within each benchmark and action architecture, these settings are kept fixed across all Stage-1 initializations.

\begin{table*}[t]
  \centering
  \footnotesize
  \setlength{\tabcolsep}{3.2pt}
  \renewcommand{\arraystretch}{1.08}
  \caption{Stage-2 training and evaluation settings for downstream VLA learning.}
  \label{tab:app:stage2:hparams}
  \begin{tabular}{@{}>{\centering\arraybackslash}p{0.12\textwidth}
                  >{\raggedright\arraybackslash}p{0.15\textwidth}
                  >{\raggedright\arraybackslash}p{0.21\textwidth}
                  >{\raggedright\arraybackslash}p{0.21\textwidth}
                  >{\raggedright\arraybackslash}p{0.21\textwidth}@{}}
    \toprule
     & \textbf{Item} & \textbf{Libero-10} & \textbf{SimplerBridge} & \textbf{RoboCasa GR1 Tabletop} \\
    \midrule
    \multirow{9}{*}{\textbf{Train}}
    & Image resolution
    & $224{\times}224$ 
    & $224{\times}224$ 
    & $224{\times}224$  \\
    & Optimizer
    & AdamW
    & AdamW
    & AdamW \\
    & AdamW betas
    & [0.9, 0.999]
    & [0.9, 0.999]
    & [0.9, 0.999] \\
    & LR
    & 1e-5
    & 5e-5
    & 5e-5 \\
    & LR schedule
    & cosine with min\_lr = 1e-6, warmup ratio 0.05
    & cosine with min\_lr = 1e-6, warmup ratio 0.05
    & cosine with min\_lr = 1e-6, warmup ratio 0.05 \\
    & Batch size
    & 128
    & 128
    & 128 \\
    & Training steps
    & 50K
    & 50K
    & 80K \\
    & Precision
    & bf16
    & bf16
    & bf16 \\
    & Action chunk size
    & 8
    & 8
    & 16 \\
    \midrule
    \multirow{4}{*}{\textbf{Eval}}
    & Action chunk size
    & 8
    & 8
    & 16 \\
    & Evaluation rollouts
    & 50 rollouts per task
    & 24 rollouts per task
    & 50 rollouts per task \\
    & Evaluation runs
    & 3 runs; mean SR
    & 3 runs; mean SR
    & 3 runs; mean SR \\
    & FM infer. steps
    & 8
    & 8
    & 8 \\
    \bottomrule
  \end{tabular}
  \renewcommand{\arraystretch}{1.0}
\end{table*}

\noindent\textbf{Robot-data pretraining.}
For robot-data pretraining in Sec.~\ref{sec:robot-data-pretrain}, we use AgiBot-World-Beta~\citep{bu2025agibot_iros} as the action-side supervision source.
Across robot-data-only, joint robot-VQA, and sequential pretraining settings, we train for 100K steps with a global batch size of 128 on 16${\times}$A800 GPUs.
The robot-pretraining action chunk size is 30.
We use learning rate $2{\times}10^{-5}$ for the VLM backbone and $1{\times}10^{-4}$ for the action head.
The action loss is weighted by 1.0.
In robot-only pretraining, the model is trained only with AgiBot action supervision.
In joint robot-VQA pretraining, the model is trained with both AgiBot action supervision and an auxiliary autoregressive VQA loss, where the loss is weighted by 0.1.
The VQA mixture includes general VQA data and multi-domain embodied VQA data.
In sequential pretraining, we first merge the \{Grounding\,+\,Egocentric Understanding\} LoRA-adapted checkpoint into the base VLM weights, and then continue LoRA-based robot-data pretraining on AgiBot-World-Beta.
Specifically, we use LoRA rank 64 and scaling factor $\alpha=128$ under pretraining, which is larger than the Stage-1 adaptation setting to allow more capacity for the stronger supervision signal from robot data.

\section{Analysis of Representation Preservation Strength}
\label{app:preservation_strength}

\noindent\textbf{Setup.}
The main experiments in Sec.~\ref{sec:axis2} compare two endpoints of representation reshaping: LoRA preserves most of the original pretrained VLM, while Full Finetune updates the whole backbone more aggressively.
To examine the intermediate regime, we vary how strongly a learned LoRA update is merged into the base VLM.
Specifically, we use the single-domain Grounding LoRA checkpoint and introduce a scalar $\lambda$ during checkpoint merging:
\[
    W_{\lambda} = W_0 + \lambda \Delta W_{\text{LoRA}},
    \qquad
    \Delta W_{\text{LoRA}} = \frac{\alpha}{r}BA .
\]
Here, $W_0$ is the original weight, $A$ and $B$ are the learned LoRA matrices, $r$ is the LoRA rank, and $\alpha$ is the LoRA scaling factor.
A value of $\lambda=0$ recovers the original pretrained VLM, $\lambda=1$ corresponds to the standard LoRA merge, and $\lambda>1$ amplifies the Stage-1 VQA update.
Changing $\lambda$ only adjusts the merge strength; the adaptation data, LoRA rank, trainable modules, and Stage-2 recipe are held fixed.
We use Grounding and Libero-10 as a controlled diagnostic setting because the main experiments show clear positive transfer in this case, making it easier to isolate the effect of update strength.

\begin{table}[h]
  \centering
  \small
  \setlength{\tabcolsep}{1pt}
  \renewcommand{\arraystretch}{1.0}
  \caption{Libero-10 success rates under different LoRA merge strengths.}
  \label{tab:app:preservation_strength}
  \begin{tabular}{@{}>{\centering\arraybackslash}p{0.14\linewidth}
                  >{\raggedright\arraybackslash}p{0.46\linewidth}
                  >{\centering\arraybackslash}p{0.28\linewidth}@{}}
    \toprule
    \textbf{$\lambda$} & \textbf{Merge setting} & \textbf{SR (\%)} \\
    \midrule
    0.0 & Original pretrained VLM (Baseline) & 92.4 \\
    0.5 & Partial LoRA merge & 94.6 \\
    1.0 & Standard LoRA merge & \textbf{95.6} \\
    1.5 & Amplified LoRA merge & 95.0 \\
    2.0 & Stronger amplified LoRA merge & 92.7 \\
    \bottomrule
  \end{tabular}
  \renewcommand{\arraystretch}{1.0}
\end{table}

\noindent\textbf{Results.}
Table~\ref{tab:app:preservation_strength} shows a non-monotonic relationship between LoRA merge strength and downstream action performance.
Increasing $\lambda$ from 0 to 0.5 and 1.0 improves the success rate from 92.4\% to 94.6\% and then to 95.6\%, indicating that the Grounding VQA update provides useful action-relevant signal when injected at an appropriate strength.
However, further amplification does not bring additional gains: the success rate decreases to 95.0\% at $\lambda=1.5$ and 92.7\% at $\lambda=2.0$.
This diagnostic is consistent with our preservation view.
It suggests that embodied VQA adaptation can help VLA initialization, but over-amplifying the update may move the model too far from the pretrained representation and weaken downstream transfer.

\section{VLM Retention Diagnostics}
\label{app:behavioral_retention}

\noindent\textbf{Setup.}
This diagnostic separates two effects of Stage-1 adaptation: how well the model learns the auxiliary embodied VQA task, and how much general VLM capability it retains.
For embodied VQA evaluation, we use a held-out validation split of 1k examples from each Stage-1 data setting.
For each example, GPT-5.4~\citep{singh2025openai} is given the question, the reference answer, and the model predicted answer, and assigns a 0--10 score based on semantic correctness.
We use the same scoring prompt and decoding setting for all compared VLMs, and report the average score.
For general VLM retention, we evaluate the same checkpoints on MMBench
~\citep{liu2024mmbench} and MMStar~\citep{chen2024we} , and report the average change from the Base VLM as $\Delta$. We also report downstream VLA average success rate to connect these diagnostics with action-learning transfer.
This diagnostic is not intended to rank all VQA domains; rather, it examines whether stronger auxiliary-task fitting under the key perception-side domains corresponds to better retention and downstream transfer.

\begin{table*}[t]
  \centering
  \footnotesize
  \setlength{\tabcolsep}{2pt}
  \renewcommand{\arraystretch}{1.08}
  \caption{Stage-1 VQA learning, general VLM retention, and downstream VLA transfer. VQA Score is the average GPT-5.4 score over 1k held-out examples. $\Delta$ is the average MMBench/MMStar change from the Base VLM. VLA Avg. SR averages Libero-10 and RoboCasa over two action heads.}
  \vspace{2mm}
  \label{tab:app:behavioral_retention}
  \begin{tabular}{@{}>{\raggedright\arraybackslash}p{0.18\linewidth}
                  >{\centering\arraybackslash}p{0.09\linewidth}
                  >{\centering\arraybackslash}p{0.13\linewidth}
                  >{\centering\arraybackslash}p{0.11\linewidth}
                  >{\centering\arraybackslash}p{0.11\linewidth}
                  >{\centering\arraybackslash}p{0.12\linewidth}
                  >{\centering\arraybackslash}p{0.12\linewidth}@{}}
    \toprule
    \textbf{VQA data} & \textbf{Update} & \textbf{VQA Score} & \textbf{MMBench} & \textbf{MMStar} & \textbf{$\Delta$} & \textbf{VLA Avg. SR} \\
    \midrule
    Base VLM & -- & -- & 84.2 & 62.4 & 0.0 & 71.4 \\
    \midrule
    Grounding & LoRA & 5.6 & 84.6 & 62.4 & \gain{0.2} & 73.2 \\
    Grounding & Full FT & 6.8 & 68.2 & 42.1 & \loss{-18.2} & 68.8 \\
    Ego Unders. & LoRA & 5.1 & 81.7 & 62.7 & \loss{-1.1} & 72.6 \\
    Ego Unders. & Full FT & 7.2 & 68.5 & 40.3 & \loss{-18.9} & 67.8 \\
    \bottomrule
  \end{tabular}
  \renewcommand{\arraystretch}{1.0}
\end{table*}

\noindent\textbf{Interpretation.}
Table~\ref{tab:app:behavioral_retention} shows that stronger auxiliary-task fitting does not necessarily produce a better VLA initialization.
In both Grounding and Egocentric Understanding, Full Finetune obtains a higher embodied VQA score than LoRA, but it loses about 18\% on average across MMBench and MMStar and falls below the Base VLM in downstream VLA average success rate.
LoRA, in contrast, keeps general VLM performance much closer to the Base VLM while still improving downstream VLA transfer.
These results provide behavioral evidence for the preservation-vs.-specialization interpretation: useful VLA initialization benefits from adding embodied signal without overly degrading the general pretrained VLM capability.

\section{In-Family Validation of Bottleneck-Aligned Transfer}
\label{app:benchmark_analysis}

\noindent\textbf{Setup.}
To further examine the bottleneck-alignment interpretation in Sec.~\ref{sec:axis1:quant}, we evaluate the same single-domain Stage-1 checkpoints on Libero-10-plus~\citep{fei2025libero}.
Libero-10-plus is harder than Libero-10, but remains within the same LIBERO-style single-arm tabletop manipulation regime.
Its increased difficulty comes from additional perturbations and noise, including changes in object layouts, camera viewpoints, robot initial states, language instructions, lighting, textures, and visual observations.
Therefore, it provides an in-family stress test that helps separate benchmark difficulty from bottleneck type.
We do not train on the Libero-10-plus training set; instead, we directly evaluate the policies trained on Libero-10 in a zero-shot manner.
This comparison allows us to examine whether the positive transfer observed on Libero-10 is specific to one benchmark split, or whether it persists when the benchmark becomes harder while retaining similar perception and manipulation bottlenecks.

\begin{table}[h]
  \centering
  \small
  \setlength{\tabcolsep}{5pt}
  \renewcommand{\arraystretch}{1.08}
  \caption{Single-domain VQA transfer on Libero-10 and Libero-10-plus. Deltas are relative to each benchmark's baseline.}
  \label{tab:app:libero_plus}
  \begin{tabular}{@{}lcc@{}}
    \toprule
    \textbf{Domain} & \textbf{Libero-10} & \textbf{Libero-10-plus} \\
    \midrule
    \rowcolor{gray!15}
    Baseline & 92.4 & 62.5 \\
    \midrule
    Spatial & 93.0\,\gain{0.6} & 64.7\,\gain{2.2} \\
    Grounding & 95.6\,\gain{3.2} & \textbf{69.5}\,\gain{7.0} \\
    Plan \& Reason. & 95.2\,\gain{2.8} & 65.2\,\gain{2.7} \\
    Camera Pred. & 93.2\,\gain{0.8} & 63.8\,\gain{1.3} \\
    Ego Unders. & 95.3\,\gain{2.9} & 69.2\,\gain{6.7} \\
    Temporal & \textbf{96.4}\,\gain{4.0} & 66.1\,\gain{3.6} \\
    Action-NTP & 95.4\,\gain{3.0} & 63.2\,\gain{0.7} \\
    \bottomrule
  \end{tabular}
  \renewcommand{\arraystretch}{1.0}
\end{table}

\noindent\textbf{Results.}
Table~\ref{tab:app:libero_plus} shows that the positive transfer also appears on Libero-10-plus.
All single-domain VQA adaptations improve over the baseline, with gains ranging from $+0.7$\% to $+7.0$\%.
This suggests that the gains on Libero-10 reflect a broader LIBERO-style transfer pattern, rather than a result specific to that particular split.
The pattern is also more consistent with the bottleneck-alignment interpretation than with a simple benchmark-difficulty explanation.
Although Libero-10-plus is harder than Libero-10, it remains within the same single-arm tabletop manipulation regime. Its additional perturbations mainly increase the demand for object localization, task planning and reasoning, and robust scene understanding, which are still closely related to the capabilities injected by embodied VQA adaptation.
Furthermore, Temporal Understanding gives the strongest result on Libero-10, whereas Grounding becomes the strongest domain on Libero-10-plus, followed closely by Egocentric Understanding.
This domain ranking shift suggests that even within the same benchmark family, the most useful injected capability depends on which downstream bottleneck becomes more prominent.

Overall, Libero-10-plus provides an in-family stress test for the results in Table~\ref{tab:axis1:domain}.
The benchmark-dependent pattern should not be understood as ``easy benchmarks benefit and hard benchmarks fail.''
Instead, these results suggest that embodied VQA adaptation is more helpful when the injected capability matches the downstream action-learning bottleneck.
When the bottleneck shifts toward real-to-sim transfer, high-dimensional control, or broader cross-scene generalization, as in SimplerBridge and RoboCasa, the same VQA adaptation may become weaker or less consistent.

\section{Additional Probing Experiments}
\label{app:probing}

\noindent\textbf{Setup.}
We conduct an additional frozen-backbone probing experiment to examine how much action-relevant information is directly available in the VLM representation.
Unlike the main initialization experiments, where the VLM backbone can be adapted during downstream action training, here we freeze the VLM backbone and train only the action head.
This setting probes whether a fixed VLM representation can directly support action decoding without further representation adaptation.
We evaluate both action-head designs using the same Stage-1 embodiment-adapted VLM as in Sec.~\ref{sec:axis1:domain}.

\begin{table*}[t]
  \centering
  \footnotesize
  \setlength{\tabcolsep}{3.2pt}
  \renewcommand{\arraystretch}{1.08}
  \caption{Frozen-backbone probing under two action heads.}
  \label{tab:app:frozen_probing}
  \begin{tabular}{@{}l ccc ccc@{}}
    \toprule
    & \multicolumn{3}{c}{MLP Head} & \multicolumn{3}{c}{Diffusion Expert} \\
    \cmidrule(r){2-4}\cmidrule(l){5-7}
    \textbf{Domain} & \textbf{Libero-10} & \textbf{SimplerBridge} & \textbf{RoboCasa} & \textbf{Libero-10} & \textbf{SimplerBridge} & \textbf{RoboCasa} \\
    \midrule
    \rowcolor{gray!15}
    Baseline & 1.0 & 0.0 & 1.0 & 43.8 & 15.0 & 15.42 \\
    \midrule
    Spatial & 1.0\,\zdelta & 0.0\,\zdelta & 1.2\,\gain{0.2} & 43.4\,\loss{-0.4} & \textbf{21.6}\,\gain{6.6} & 14.08\,\loss{-1.34} \\
    Grounding & 1.0\,\zdelta & \textbf{0.3}\,\gain{0.3} & 1.4\,\gain{0.4} & \textbf{55.2}\,\gain{11.4} & 19.0\,\gain{4.0} & \textbf{17.17}\,\gain{1.75} \\
    Plan \& Reason. & 0.7\,\loss{-0.3} & 0.0\,\zdelta & 0.8\,\loss{-0.2} & 44.3\,\gain{0.5} & 14.6\,\loss{-0.4} & 15.50\,\gain{0.08} \\
    Camera Pred. & 0.5\,\loss{-0.5} & 0.0\,\zdelta & 0.3\,\loss{-0.7} & 44.6\,\gain{0.8} & 11.5\,\loss{-3.5} & 14.50\,\loss{-0.92} \\
    Ego Unders. & 1.1\,\gain{0.1} & 0.1\,\gain{0.1} & 0.7\,\loss{-0.3} & 52.4\,\gain{8.6} & 18.6\,\gain{3.6} & 15.51\,\gain{0.09} \\
    Temporal & 0.9\,\loss{-0.1} & 0.1\,\gain{0.1} & 1.1\,\gain{0.1} & 42.0\,\loss{-1.8} & 8.0\,\loss{-7.0} & 16.42\,\gain{1.00} \\
    Action-NTP & \textbf{1.4}\,\gain{0.4} & \textbf{0.3}\,\gain{0.3} & \textbf{1.5}\,\gain{0.5} & 48.2\,\gain{4.4} & 17.2\,\gain{2.2} & 16.05\,\gain{0.63} \\
    \bottomrule
  \end{tabular}
  \renewcommand{\arraystretch}{1.0}
\end{table*}

\noindent\textbf{Results.}
The MLP-head results in Table~\ref{tab:app:frozen_probing} show that the frozen VLM representation is almost unusable with the lightweight MLP head.
Across all domains and benchmarks, success rates remain in the 0--2\% range.
This suggests that a low-capacity action head cannot directly decode effective actions from fixed VLM features, even when the VLM has been adapted with embodied VQA supervision.
However, the results change with the higher-capacity diffusion expert.
As shown in Table~\ref{tab:app:frozen_probing} (diffusion-expert), frozen VLM representations begin to support continuous action decoding when paired with a stronger action head, although the performance remains limited. The performance also varies with the capability injected by each embodied VQA domain.
Grounding gives the strongest frozen-backbone probing result on Libero-10 and RoboCasa, while Spatial achieves the best result on SimplerBridge.
Beyond this, Egocentric Understanding and Action-NTP are two other robust positive domains, as each improves over the frozen baseline on most benchmarks with the diffusion expert.
This pattern is consistent with the main single-domain results in Sec.~\ref{sec:axis1:domain}, where Grounding, Egocentric Understanding, and Action-NTP provide relatively reliable transfer signals.
It further supports the interpretation that these capabilities inject action-relevant information into the VLM representation.

Overall, this probing experiment complements our main initialization results.
It suggests that embodied VQA adaptation can make some frozen VLM representations more action-decodable, especially when the injected capability is related to grounding, egocentric understanding, or action structure.
At the same time, the large gap between frozen probing and the main training results in Sec.~\ref{sec:axis1:domain} indicates that effective VLA initialization is not simply a matter of directly decoding actions from static VLM features.
Useful initialization requires VLM representations that can be further adapted through an appropriate action-learning pipeline.

\section{Limitations and Future Work}
\label{app:limitations}

$\bullet$~\textbf{Simulation-centered evaluation.}
Our experiments are conducted on simulated manipulation benchmarks, which enable controlled and reproducible comparisons across many initialization settings.
However, they cannot fully capture real-world deployment factors such as hardware variation, sensing noise, calibration errors, and dynamics mismatch.
Future work should validate whether the bottleneck-alignment and representation-preservation patterns observed in this study persist on physical robot platforms.

$\bullet$~\textbf{Coarse VQA taxonomy and data quality.}
We organize embodied VQA data into seven capability-oriented domains, but this taxonomy is not exhaustive.
Different prompt formats, data-quality filters, finer-grained domain definitions, or even sampling ratios may change the transfer behavior.
Our next step is to develop quality-aware and bottleneck-aware data selection methods that choose Stage-1 supervision according to the target action-learning setting.

$\bullet$~\textbf{Limited robot-data pretraining scale.}
Our robot-data pretraining analysis uses a single robot-data source and focuses on a RoboCasa evaluation setting, which keeps the comparison controlled but limits conclusions about broader robot pretraining.
As robot pretraining data becomes broader and more diverse, the balance between action specialization and pretrained representation preservation may become different.
Therefore, we will explore whether the staged, adapter-based pattern persists under larger and more heterogeneous robot-data regimes in the future.

$\bullet$~\textbf{Mechanistic understanding of representation preservation.}
Our preservation interpretation is supported by downstream transfer results, merge-strength analysis, behavioral retention diagnostics, and frozen-backbone probing, but the analysis remains primarily empirical.
Future work should use feature-level diagnostics or causal interventions to identify which visual-language features are preserved, overwritten, or converted into action-relevant representations during VLM-to-VLA adaptation.

\newpage
\end{document}